\documentclass{article} %
\usepackage{iclr2027_conference,times}

\usepackage{amsmath,amsfonts,bm}

\def\eqref#1{equation~\ref{#1}}

\def\1{\bm{1}}

\DeclareMathAlphabet{\mathsfit}{\encodingdefault}{\sfdefault}{m}{sl}
\SetMathAlphabet{\mathsfit}{bold}{\encodingdefault}{\sfdefault}{bx}{n}

\usepackage{hyperref}
\hypersetup{hidelinks}
\usepackage{url}
\usepackage[T1]{fontenc}
\usepackage{tgtermes}
\usepackage{graphicx}
\usepackage{longtable,booktabs,tabularx,array, multirow}
\usepackage{pdflscape}
\usepackage{xcolor}
\usepackage{caption}
\usepackage{enumitem}
\usepackage[table]{xcolor}
\usepackage{wrapfig}  
\usepackage{amsmath} 

\let\arenafilledcite\cite
\renewcommand{\cite}[1]{\if\relax\detokenize{#1}\relax\mbox{\textbf{?}}\else\arenafilledcite{#1}\fi}
\definecolor{FrontierRule}{HTML}{C6CDD2}

\usepackage{tikz}
\usepackage{booktabs}
\usepackage{array}
\usepackage{xcolor}
\usepackage{placeins}

\usetikzlibrary{
  arrows.meta,
  positioning,
  calc
}

\definecolor{ReuseGreen}{HTML}{087F5B}
\definecolor{ReuseGreenMid}{HTML}{9FD4C1}
\definecolor{ReuseGreenLight}{HTML}{EDF7F3}
\definecolor{ReuseBlue}{HTML}{3979A8}
\definecolor{ReuseOrange}{HTML}{C96A18}
\definecolor{ReuseGray}{HTML}{59636E}
\definecolor{ReuseLightGray}{HTML}{DDE3E7}

\title{Auto-Research Arena: Benchmarking and Advancing Auto-Research in Language Modeling}

\author{%
\textbf{Xi Chen$^{1}$}\quad
\textbf{Chongjian Ge$^{2}$}\quad
\textbf{Yiran Xu$^{2}$}\quad
\textbf{Yiqun Mei$^{2}$}\quad
\textbf{Henrik I. Christensen$^{1}$}\\
\textbf{Hao Tan$^{2}$}\quad
\textbf{Ziwen Chen$^{2,\dagger}$}\quad
\textbf{Hanwen Jiang$^{2,\dagger}$}\\
$^{1}$University of California San Diego \quad
$^{2}$Adobe Research
}

\newcommand{\methodname}{Auto-Research Arena}
\newcommand{\shortname}{Arena}

\iclrfinalcopy %
\begin{document}

\maketitle
{\renewcommand{\thefootnote}{\fnsymbol{footnote}} \footnotetext[2]{Equal Supervision.}

\begin{abstract}
As autonomous research systems become more capable, benchmarks must both
evaluate their performance and support meaningful research progress.
We introduce \methodname{}, a multi-target benchmark for evaluating autonomous
research systems and discovering ideas in language modeling.
Arena supports diverse research objectives within a shared experimental
environment by separating \emph{admissibility} from \emph{improvement}: model
quality and resource constraints determine whether a result qualifies, while
the selected target measures progress among admissible results.
We instantiate Arena in language model pretraining with nine targets spanning
model, data, training, and inference efficiency, and evaluate seven research
harnesses using two foundation models.
Arena clearly differentiates existing harnesses across research objectives,
with several sophisticated methods outperformed by our simple sequential search
baseline, indicating substantial headroom in current auto-research systems.
Beyond evaluation, Arena reveals a substantially broader frontier of model and
system ideas aligned with active research directions.
Reusing these discoveries in a new quality-driven search yields lower training
loss under matched data and FLOPs and lower held-out BPB at the same model size.
Together, Arena provides a common framework for benchmarking autonomous
research and expanding its scope in language modeling. The code is available at \href{https://github.com/Auto-Research-Arena/auto-research-arena}{https://github.com/Auto-Research-Arena/auto-research-arena}.
\end{abstract}

\begin{figure*}[h]
    \centering
    \includegraphics[width=\textwidth]{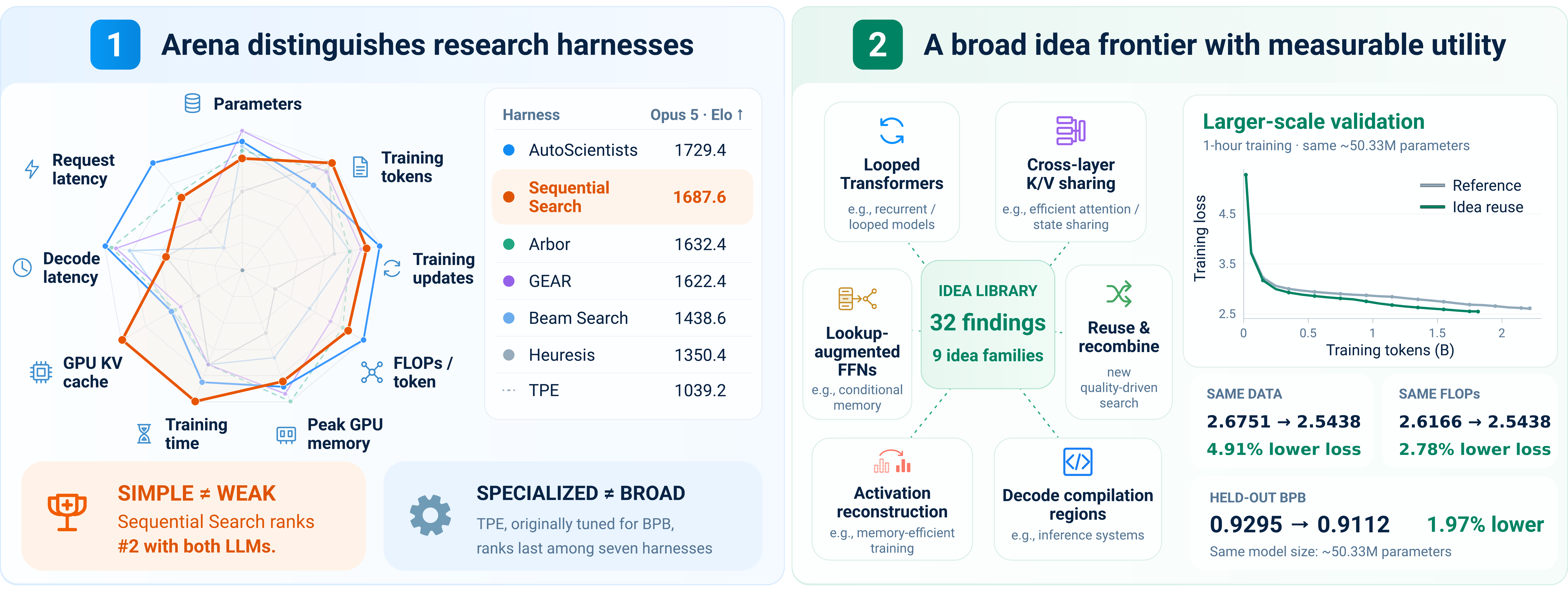}
    \vspace{-0.3in}
    \caption{\small{
\textbf{Auto-Research Arena evaluates research systems and expands the idea frontier.}
\textbf{Left:} Arena differentiates research harnesses across nine complementary objectives. Our simple Sequential Search baseline remains competitive, whereas the TPE method tailored to validation-loss minimization underperforms in this broader evaluation.
\textbf{Right:} Multi-target research reveals a broad frontier of ideas. A program combining these discoveries achieves lower training loss than the reference program under matched data and FLOP budgets, as well as lower final held-out BPB at the same model size.
}}
    \label{fig:teaser}
    \vspace{-0.15in}
\end{figure*}

\vspace{-1mm}
\section{Introduction}
\label{sec:intro}
\vspace{-2mm}

Advances in LLMs and research harnesses have enabled increasingly capable autonomous research systems~\citep{AI-Researcher,AutoResearchClaw}.
These systems iteratively propose ideas, conduct experiments, and use the results to guide further research.
Auto-research benchmarks provide a common basis for comparing these systems, but their role extends beyond evaluation.
By defining clear objectives and constraints, a well-designed benchmark helps researchers identify, validate, and build on useful improvements.
However, existing benchmarks remain limited both in evaluating research systems and in supporting broad research ideas~\citep{MLE-Bench,RE-Bench,MLGym,karpathyautoresearch,NanoGPT-Bench}.

\vspace{-0.5mm}
To address this gap, we introduce \textbf{\methodname{}}, a comprehensive benchmark for evaluating auto-research systems and advancing idea discovery in language modeling.
Departing from prior benchmarks that define progress primarily through a particular objective within each research problem, \textbf{Arena defines multiple research objectives within a common experimental environment}.
These objectives address complementary bottlenecks across the model lifecycle, including model size, data consumption, training computation and memory, and inference memory and latency.
The shared environment provides a common experimental basis for evaluating whether a harness remains effective across different research objectives.
Meanwhile, different objectives reward different types of improvements.
For example, reducing model size or inference memory can count as progress even when prediction loss remains unchanged.
By recognizing these complementary forms of progress, Arena supports exploration of a broader range of model designs and systems optimizations.

\vspace{-0.5mm}
To support this broad range of research objectives within a unified framework, Arena specifies both what to improve and which requirements a candidate must satisfy. That is, Arena separates the standard for \emph{admissibility} from the direction of \emph{improvement}.
Each objective is formulated as a constrained optimization problem.
Constraints on model quality and relevant non-target resources determine whether a candidate is \emph{admissible}.
Among admissible candidates, a reduction in the selected resource metric constitutes \emph{improvement}.
Changing the target metric defines a different research objective within the same framework.
This formulation supports consistent evaluation across diverse objectives, while excluding gains that violate the specified quality or resource constraints.
\vspace{-0.5mm}
Beyond evaluating objectives independently, Arena supports research across objectives.
Separate searches can pursue different objectives in parallel, each exploring improvements relevant to its target.
Compatible improvements discovered in one search can then be incorporated into the starting program for a subsequent search targeting a different objective.
This enables both \emph{parallel exploration} of distinct research directions and \emph{cumulative progress} through the reuse of earlier discoveries.
By explicitly specifying objectives, constraints, budgets, and starting programs, Arena makes it possible to measure how prior discoveries affect subsequent research.

\vspace{-0.5mm}
We evaluate six LLM-based auto-research methods and a Tree-structured Parzen Estimator baseline~\citep{TPE} across Arena's nine objectives. \textbf{Arena provides a discriminative and meaningful evaluation of
research harnesses}: methods exhibit clear differences across objectives, with no single harness dominating the benchmark. Two results further support the evaluation. 
TPE, specialized for the original validation loss-oriented setting, performs worst in Arena's broader multi-target evaluation. 
Moreover, our simple Sequential Search baseline ranks second overall with both foundation models, outperforming several substantially more sophisticated harnesses and suggesting large headroom in current auto-research system design.
The recorded search trajectories help explain how these final outcomes arise.
Higher-performing harnesses tend to sustain progress throughout the search budget, whereas lower-performing methods often plateau earlier or improve less consistently.

\vspace{-0.5mm}
\textbf{Multi-target research also exposes a substantially broader idea
frontier}. 
We empirically find that different objectives lead to qualitatively different discoveries, yielding 32 findings across nine idea families spanning model architecture, optimization, memory, and inference systems.
Many of these discoveries align with active directions in LLM research.
For example, when optimizing parameter efficiency, agents discover looped Transformers through weight sharing.
They further learn depth-specific residual gains to adapt the residual updates across repeated applications of the shared block.
This finding closely parallels recent work identifying residual adaptation as a key ingredient of looped Transformers~\citep{wang2026residual}.
Similar connections appear across other discoveries, although alignment with existing research does not imply identical solutions.
Several discoveries follow design principles similar to those of prior work but implement them through distinct mechanisms, offering alternative approaches to the same research problems.
Together, these findings show how Arena's diverse objectives expand the range of ideas explored by autonomous research systems.

Finally, we examine whether ideas discovered through Arena's lightweight searches remain useful beyond the search setting.
The original searches train each candidate for only 10 minutes.
We provide the collected discoveries to a new search focused on model quality, which can select, combine, and modify these ideas.
We then train the resulting model for one hour from random initialization.
Unlike the original searches, which optimize individual efficiency targets, this experiment evaluates improvements in model quality directly.
Relative to the reference run, training loss decreases from 2.675 to 2.544 at matched training data and from 2.617 to 2.544 at matched cumulative FLOPs.
Held-out BPB also decreases from 0.9295 to 0.9112 at essentially the same model size.
These results show that discoveries from Arena's small-scale, multi-target searches can improve model quality through combination and adaptation.
Arena thus supports both the evaluation of autonomous research systems and the discovery of reusable research ideas.

\vspace{-1mm}
\section{Related Work}
\vspace{-2mm}

\noindent \textbf{Autonomous research benchmarks.}
PaperBench~\citep{PaperBench} evaluates replication of published research, while EXP-Bench~\citep{EXP-Bench} assesses the design and execution of paper-derived experiments. These benchmarks emphasize implementing and validating specified research content.
More open-ended settings include MLE-bench~\citep{MLE-Bench} and
RE-Bench~\citep{RE-Bench}, which evaluate ML engineering and research, alongside MLGym~\citep{MLGym}, MLRC-Bench~\citep{MLRC-Bench}, and AstaBench~\citep{AstaBench}, which support standardized experimentation and evaluation across ML and scientific tasks.
Within language modeling, Karpathy's autoresearch~\citep{karpathyautoresearch} minimizes validation loss under a fixed training-time budget, whereas NanoGPT Bench~\citep{NanoGPT-Bench} targets faster training at a required model quality.
Related benchmarks evaluate reproduction of NanoGPT improvements~\citep{Automated-LLM-Speedrunning} and correctness-constrained computational speedups~\citep{KernelBench,AlgoTune,GSO}.
However, existing benchmarks provide limited evaluation of auto-research systems across complementary objectives. \shortname{} instead unifies diverse objectives within a shared experimental setting, revealing how the research objective shapes both harness performance and the resulting idea frontier.

\noindent \textbf{Autonomous research systems.}
Complementing these benchmarks, research harnesses determine how agents organize experiments and use accumulated evidence. End-to-end systems connect ideation, experimentation, and reporting~\citep{lu2024aiscientist,AI-Researcher}, while AIDE~\citep{AIDE} and AIRA~\citep{AIRA} organize iterative code improvement through search. AutoScientists~\citep{AutoScientists} coordinates decentralized teams, Arbor~\citep{Arbor} maintains a persistent hypothesis tree, and GEAR~\citep{GEAR} evolves populations of research states. Heuresis~\citep{Heuresis} further compares search strategies across quality, diversity, and novelty. 
\shortname{} evaluates how the relative strengths of research harnesses vary across objectives.

\vspace{-1mm}
\section{Auto-Research Arena}
\label{sec:arena}
\vspace{-2mm}

Auto-Research Arena defines a family of research problems that optimize different resource targets within a shared language-modeling environment. 
Fig.~\ref{fig:overview} illustrates the research loop, from selecting a target and evaluating candidate programs to identifying each harness's best admissible program and comparing performance across targets.
In the following, we present the general formulation in Sec.~\ref{sec:formulation} and its language-modeling instantiation in Sec.~\ref{sec:lm-instantiation}.

\vspace{-1mm}
\subsection{Auto-Research as Constrained Optimization}
\label{sec:formulation}
\vspace{-1mm}

Consider a testbed with a fixed data pool, evaluator, and execution environment.
An auto-research harness searches over executable programs $p \in \mathcal{P}$, where candidate programs may modify the model architecture, optimizer, training schedule, and inference implementation.

For each objective $j$, Arena minimizes a resource metric $\operatorname{Target}_j(p)$ subject to a quality threshold $\tau$ (shared by all targets $j$), target-specific resource limits $\mathbf{b}_j$, and correctness checks:
\vspace{0.03in}
{\setlength{\abovedisplayskip}{4pt}
 \setlength{\belowdisplayskip}{4pt}
\begin{equation}
\small
\setlength{\jot}{1pt}
\begin{aligned}
    \underset{p \in \mathcal{P}}{\operatorname{minimize}}
    \quad & \operatorname{Target}_j(p) \\
    \operatorname{subject~to}
    \quad & \mathcal{L}(p) < \tau, \\
    & \operatorname{Resources}_j(p) \leq \mathbf{b}_j .
\end{aligned}
\label{eq:arena}
\end{equation}
}
Here, $\mathcal{L}(p)$ denotes validation loss, for which lower values are better.
The vector $\operatorname{Resources}_j(p)$ contains the constrained resource measurements, and $\mathbf{b}_j$ specifies their limits.
Resource inequalities apply componentwise.
This formulation separates \emph{admissibility} from \emph{improvement}: candidates qualify by satisfying the quality, resource, and correctness requirements, while lower target values indicate better outcomes among qualifying candidates.

\vspace{-1mm}
\noindent \textbf{Efficiency targets.}
Arena instantiates this formulation with nine objectives spanning model, training, and inference efficiency.
These objectives address complementary bottlenecks in model development and deployment.
Tab.~\ref{tab:efficiency-targets} summarizes their measurements and practical motivations.

Each metric captures a distinct dimension of efficiency.
Parameter efficiency measures the model size needed to achieve the required quality.
Training efficiency measures the resources needed to reach that quality, while inference efficiency measures the resources needed to use the trained model.
Optimizing these metrics separately defines distinct research problems within the same experimental environment.

\begin{figure}[t]
  \centering
  \includegraphics[width=\linewidth]{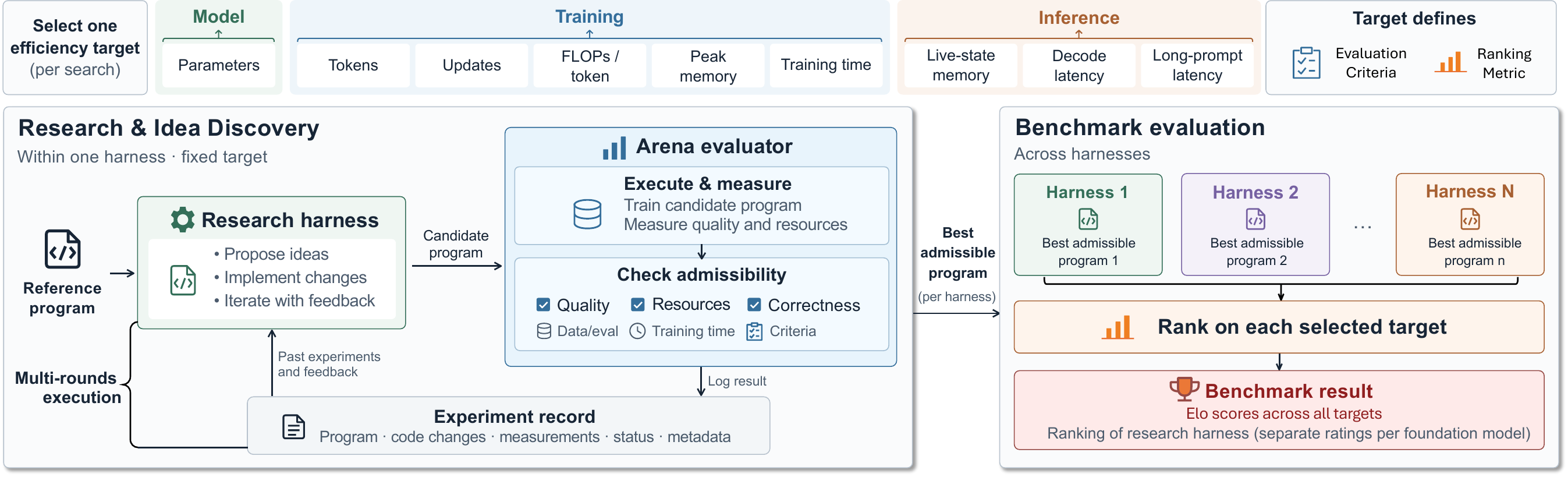}
  \vspace{-0.27in}
  \caption{\small{\textbf{Overview of Auto-Research Arena Pipeline.}
  Each search run optimizes one of nine efficiency targets.
  A research harness proposes and implements changes to a reference program.
  Arena evaluates the resulting candidates, records outcomes, and provides feedback.
  The best admissible candidate determines the benchmark result.}}
  \label{fig:overview}
  \vspace{-0.25in}
\end{figure}

\vspace{-2mm}
\subsection{Language-Modeling Instantiation}
\label{sec:lm-instantiation}
\vspace{-1mm}

We instantiate Arena using Karpathy's autoresearch language-modeling testbed~\citep{karpathyautoresearch}, in which candidate solutions are compact LLM pretraining programs.
We retain its fixed training-data pool and lightweight experimental setup, replacing loss minimization with the constrained formulation in Eq.~\ref{eq:arena}.
Each track optimizes one of nine efficiency metrics subject to requirements on model quality and relevant non-target resources.

\noindent\textbf{Model quality.}
Following the original testbed~\citep{karpathyautoresearch}, we measure language-modeling quality using validation bits per byte (BPB).
A candidate is admissible only if its validation BPB is below $1.05$.
This threshold is slightly above the reference program's BPB, providing flexibility for structural changes while limiting quality degradation.

\noindent\textbf{Resource constraints.}
Each track limits relevant non-target resources to control the trade-offs permitted during optimization.
These constraints cover quantities such as model size, training FLOPs, peak training memory, and available training data.
Inference tracks also constrain other serving costs.
For example, the live-state-memory track imposes a latency bound, while the decode-latency track imposes a live-state-memory bound.
Each track therefore measures improvement in its selected metric while requiring other resource costs to remain within specified limits.

\noindent\textbf{Measurement.}
The Arena evaluator measures all target metrics and constrained resources directly.
Depending on the target, measurements are taken after a fixed 600-second training period or when the candidate first meets the quality requirement.
Training metrics include consumed tokens, FLOPs per token, peak allocated GPU memory, elapsed training time, and optimizer updates.
Inference metrics are measured after training using standardized prompt and generation workloads.
Appendix~\ref{app:target-definitions} provides complete metric definitions and resource constraints for all nine tracks.

\begin{wrapfigure}[8]{r}{0.48\textwidth}
\scriptsize
\setlength{\abovedisplayskip}{2pt}
\setlength{\belowdisplayskip}{2pt}
\setlength{\jot}{1pt}
\vspace{-0.15in}
\begin{equation*}
\begin{aligned}
    \underset{p}{\operatorname{minimize}}
    \quad & N_{\mathrm{param}}(p) \\
    \text{subject to}
    \quad & \operatorname{BPB}(p) < 1.05,\\
    & \operatorname{FLOPs}(p) \leq 239.1\,\text{M/token},\\
    & \operatorname{Mem}_{\mathrm{train}}(p) \leq 44.0\,\text{GiB}.
\end{aligned}
\end{equation*}
{\scriptsize\raggedright
The BPB threshold of $1.05$ permits a modest quality decrease relative to the
reference program to accommodate structural changes.
The compute and memory limits match the reference program's measured costs,
ensuring that parameter reductions do not exceed either resource budget.
\par}
\end{wrapfigure}

\begin{table}[t]
\centering
\scriptsize
\caption{\small{\textbf{Efficiency targets in Auto-Research Arena.}
Each track minimizes one target subject to a shared model-quality
threshold and target-specific resource constraints.}}
\vspace{-0.1in}
\label{tab:efficiency-targets}
\setlength{\tabcolsep}{4pt}
\renewcommand{\arraystretch}{1.12}
\begin{tabularx}{\linewidth}{@{}
  >{\raggedright\arraybackslash\leftskip=0.7em}p{0.18\linewidth}
  >{\raggedright\arraybackslash}p{0.32\linewidth}
  >{\raggedright\arraybackslash}X@{}}
\toprule
\multicolumn{1}{@{}l}{\textbf{Target}} &
\textbf{Measured quantity} & \textbf{Why it matters} \\
\midrule

\multicolumn{3}{@{}l}{\textbf{Model}} \\
Total parameters &
Number of model parameters &
Model storage and weight-memory footprint~\citep{lan2019albert}. \\

\addlinespace[4pt]
\multicolumn{3}{@{}l}{\textbf{Training}} \\
Data consumption &
Tokens used to reach target quality &
Learning efficiency under a token budget~\citep{hoffmann2022computeoptimal}. \\
\addlinespace[2pt]
Optimizer updates &
Updates to reach the quality threshold &
Convergence in fewer updates~\citep{gupta2018shampoo}. \\
\addlinespace[2pt]
Training FLOPs &
FLOPs per training token &
Compute demand per token~\citep{hoffmann2022computeoptimal}. \\
\addlinespace[2pt]
Peak memory &
Peak allocated GPU memory &
Training feasibility on memory limits~\citep{chen2016sublinear}. \\
\addlinespace[2pt]
Training time &
Time to reach the quality threshold &
Training turnaround~\citep{mattson2020mlperf}. \\

\addlinespace[4pt]
\multicolumn{3}{@{}l}{\textbf{Inference}} \\
Live-state memory &
Memory per active request &
Concurrent serving capacity~\citep{kwon2023pagedattention}. \\
\addlinespace[2pt]
Decode latency &
Sequential decoding latency &
Autoregressive generation latency~\citep{mqa}. \\
\addlinespace[2pt]
Long-prompt latency &
Prefill + decoding latency &
Long-context response latency~\citep{zhong2024distserve}. \\
\bottomrule
\end{tabularx}
\vspace{-0.1in}
\end{table}

\noindent \textbf{Example: parameter efficiency.}
The parameter-efficiency track seeks the smallest model that reaches the required quality within the specified training-compute and memory budgets.
The remaining eight tracks follow the same formulation, each with its own target metric and corresponding resource constraints.

\vspace{-2mm}
\section{Experiments}
\label{sec:experiments}
\vspace{-2mm}

We evaluate \shortname{} along its two central roles: benchmarking autonomous
research systems and supporting broad idea discovery.
We then use its controlled multi-target setting to study how research evolves
under different objectives, constraints, budgets, and research histories.

\vspace{-1mm}
\noindent\textbf{Benchmark--harness interface.}
Arena provides the environment, task specification, reference program,
evaluator, and experiment budget, while each harness determines how to conduct
the search.
Starting from the reference program, a harness proposes code modifications,
submits candidates for evaluation, observes the measured outcomes, and decides
what to explore next.
Arena records every experiment and independently reports the best recorded
candidate satisfying the quality, resource, and correctness requirements,
regardless of which program the harness ultimately retains.
Each search receives 100 charged experiments; an experiment is charged once GPU
execution begins, including runs that subsequently fail or are abandoned.
Appendix~\ref{app:reference-baselines} provides the full interface and
protocol.

\vspace{-1mm}
\noindent\textbf{Reference baselines.}
We introduce two simple reference harnesses that differ primarily in how many
research directions they retain.
\emph{Sequential Search} maintains one program and evaluates one modification
at a time, adopting an admissible improvement before continuing.
\emph{Beam Search} retains up to four programs and evaluates two modifications
of each before selecting the next frontier.
Both condition subsequent proposals on the accumulated experimental history,
including unsuccessful attempts.
Appendix~\ref{app:reference-baselines} provides their detailed proposal,
selection, and stopping policies.

\vspace{-1mm}
\noindent\textbf{Tasks and resources.}
We evaluate all nine Arena objectives defined in Sec.~\ref{sec:arena}, each
with target-specific resource constraints and the shared quality requirement
BPB $<1.05$.
Each candidate runs on one NVIDIA A100 80GB GPU with seed 42 and up to
600 seconds of training, excluding compilation, validation, and the first ten
warmup updates.
All searches use the same approximately 631M-token ClimbMix~\citep{ClimbMix}
training pool and held-out BPB evaluator; training-token consumption includes
repeated exposure to the data.

\vspace{-1mm}
\noindent\textbf{Models and harnesses.}
We evaluate six LLM-based research harnesses with Claude Opus 5~\citep{opus5}
and GPT-6 Astra~\citep{gpt-astra}, both using medium reasoning effort.
In addition to Sequential Search and Beam Search, we evaluate
AutoScientists~\citep{AutoScientists}, Arbor~\citep{Arbor},
GEAR~\citep{GEAR}, and Heuresis~\citep{Heuresis}, which respectively use
multi-agent coordination, hypothesis trees, evolutionary search, and
diversity-preserving archives.
We additionally include TPE~\citep{TPE}, a non-agentic baseline restricted to a
predefined hyperparameter space, in the Opus 5 experiments.

\vspace{-1mm}
\noindent\textbf{Scoring.}
Each objective is scored by the best recorded admissible target value, with
lower values better.
For training-token consumption, optimizer updates, and training time, we use the
first 30-second validation checkpoint satisfying BPB $<1.05$.
We summarize performance across objectives with different units and scales using
Elo ratings from pairwise harness comparisons within each objective.
Appendix~\ref{app:elo} gives the scoring procedure.

\vspace{-1.5mm}
\subsection{Benchmarking Auto-Research Harnesses}
\label{sec:benchmark-results}
\vspace{-1.5mm}

Tab.~\ref{tab:main-results} reports the best results across Arena's nine
objectives, while Fig.~\ref{fig:research-progress} shows representative
search trajectories.

\noindent\textbf{Arena provides a discriminative evaluation.}
Research harnesses exhibit substantial differences across Arena's objectives,
with no single method dominating the benchmark.
With Opus~5, AutoScientists achieves the highest aggregate Elo, followed by
Sequential Search and Arbor; with GPT-6, GEAR ranks first, followed by
Sequential Search.
The variation across objectives shows that Arena distinguishes complementary
strengths of research systems rather than reducing their performance to a
single research setting.

\noindent\textbf{Simple search is surprisingly competitive.}
Despite its minimal design, Sequential Search ranks second overall with both
foundation LLMs, outperforming several substantially more sophisticated
harnesses.
Its strong performance suggests substantial headroom in current auto-research
system design: more elaborate orchestration does not yet consistently translate
into stronger performance.

\noindent\textbf{Specialization does not necessarily transfer across research problems.}
TPE provides a complementary sanity check.
Designed for Karpathy's autoresearch setting~\citep{TPE}, TPE performs
substantially worse than all LLM-based harnesses across Arena's broader set of
objectives.
Strong performance in a specialized research setting therefore need not
translate into broad research capability, illustrating the value of evaluating
research systems across complementary objectives.

\newsavebox{\arenamainresultstable}
\begin{table}[t]
  \centering
  \footnotesize
  \setlength{\tabcolsep}{2.5pt}
  \renewcommand{\arraystretch}{1.12}
  \caption{\small{\textbf{Benchmark performance across nine research objectives.}
Lower objective values and higher Elo ratings are better; bold indicates the
best result within each foundation-model group.
\emph{Charged experiments} reports the mean number of budgeted experiments
across objectives.
Elo is computed separately within each foundation-model group.
TPE is a non-agentic baseline and is included only in the Opus~5 comparison.}}
  \vspace{-0.1in}
  \label{tab:main-results}
  \global\setbox\arenamainresultstable=\hbox{%
  \begin{tabular}{@{}cl*{11}{c}@{}}
  \toprule
  \multirow{3}{*}{\shortstack{\textbf{Agent}\\\textbf{model}}}
    & \multicolumn{1}{c}{\multirow{3}{*}{\textbf{Harness}}}
    & \multicolumn{1}{c}{\textbf{Model}}
    & \multicolumn{5}{c}{\textbf{Training efficiency}}
    & \multicolumn{3}{c}{\textbf{Inference efficiency}}
    & \multirow{3}{*}{\shortstack{\textbf{Charged}\\\textbf{experiments}\\\textbf{(mean)}}}
    & \multirow{3}{*}{\textbf{Elo} $\uparrow$} \\
  \cmidrule(lr){3-3}
  \cmidrule(lr){4-8}
  \cmidrule(lr){9-11}
    & & Params. & Tokens & Updates & Time & FLOPs & Memory
    & KV cache & Decode & Request & & \\
    & & (M) & (M) & & (s) & (M/token) & (GiB)
    & (KiB) & (ms) & (ms) & & \\
  \midrule
      \multicolumn{2}{@{}l}{\textbf{Reference starting point}}
      & 50.33 & 398.98 & 761 & 600 & 239.08 & 43.957
      & 33,790 & 465.28 & 641.60 & -- & --\\
    \midrule

    \multirow{7}{*}{Opus 5}
      & AutoScientists~\citep{AutoScientists}
      & 13.91 & 184.55 & \textbf{267} & 240
      & \textbf{52.15} & 0.521 & 58.00
      & \textbf{49.70} & \textbf{29.46} & 84.7 & \textbf{1729.4} \\
      & Sequential Search
      & 15.15 & \textbf{152.46} & 295 & \textbf{210}
      & 56.66 & 0.567 & \textbf{1.000}
      & 98.56 & 61.30 & 87.7 & 1687.6 \\
      & Arbor~\citep{Arbor}
      & 14.55 & 159.78 & 337 & 270
      & 58.30 & \textbf{0.373} & 94.50
      & 57.09 & 57.88 & 54.0 & 1632.4 \\
      & GEAR~\citep{GEAR}
      & \textbf{13.20} & 164.43 & 306 & 270
      & 62.23 & 0.434 & 121.5
      & 56.01 & 97.84 & 100.0 & 1622.4 \\
      & Beam Search
      & 14.22 & 197.66 & 335 & 270
      & 69.85 & 0.951 & 68.00
      & 65.15 & 180.56 & 96.9 & 1438.6 \\
      & Heuresis~\citep{Heuresis}
      & 17.83 & 156.17 & 377 & 270
      & 84.15 & 1.582 & 488.5
      & 97.82 & 129.03 & 100.0 & 1350.4 \\
      & TPE~\citep{TPE}
      & 26.35 & 387.71 & 753 & 510
      & 100.27 & 5.949 & 17,790
      & 230.74 & 295.39 & 100.0 & 1039.2 \\

    \midrule

    \multirow{6}{*}{GPT-6}
      & AutoScientists~\citep{AutoScientists}
      & 19.84 & 311.16 & 508 & 420
      & 103.81 & 6.006 & 423.5
      & 177.78 & 472.56 & 30.2 & 1298.7 \\
      & Sequential Search
      & \textbf{15.28} & 223.61 & 350 & 330
      & 74.54 & \textbf{0.646} & 362.0
      & 72.39 & 116.05 & 100.0 & 1702.4 \\
      & Arbor~\citep{Arbor}
      & 18.87 & 263.19 & 405 & 480
      & 77.71 & 1.317 & 3177
      & 137.75 & 163.25 & 14.7 & 1459.6 \\
      & GEAR~\citep{GEAR}
      & 18.61 & \textbf{159.91} & \textbf{302} & \textbf{240}
      & \textbf{67.25} & 0.709 & \textbf{51.51}
      & \textbf{62.74} & \textbf{47.34} & 100.0 & \textbf{1835.3} \\
      & Beam Search
      & 27.26 & 274.46 & 339 & 480
      & 81.00 & 3.914 & 272.5
      & 142.69 & 405.06 & 95.8 & 1423.3 \\
      & Heuresis~\citep{Heuresis}
      & 33.03 & 264.37 & 355 & 420
      & 125.24 & 2.318 & 8193
      & 317.78 & 503.57 & 86.8 & 1280.6 \\

    \bottomrule
  \end{tabular}%
  }
  \resizebox{\linewidth}{!}{\usebox{\arenamainresultstable}}
  \vspace{-0.1in}
\end{table}

\begin{figure}[t]
  \centering
  \includegraphics[width=\linewidth]{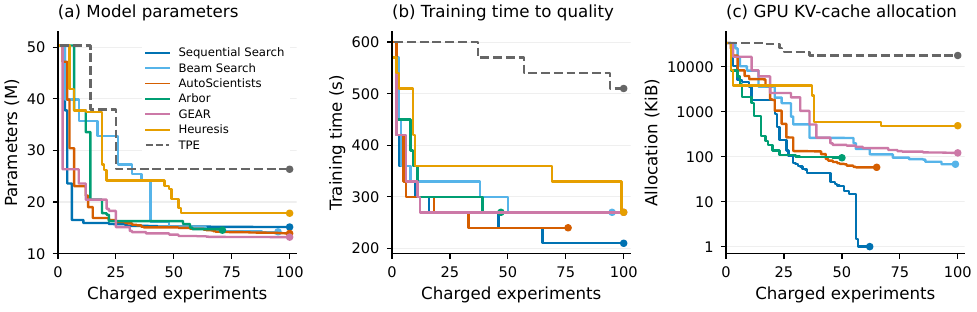}
  \vspace{-0.3in}
  \caption{\small{\textbf{Research trajectories behind the Opus~5 benchmark results.}
Curves track the best admissible result found as each search proceeds, ordered
by charged experiments; lower is better.
Counts include the reference evaluation and charged failures, and dots mark the
end of each recorded search.
Training time is measured at 30-second validation checkpoints, and KV-cache
allocation is shown on a logarithmic scale.}}
  \label{fig:research-progress}
  \vspace{-0.2in}
\end{figure}

\noindent\textbf{Research trajectories support the benchmark scores.}
The endpoint differences in Tab.~\ref{tab:main-results} are also reflected in
the search trajectories in Fig.~\ref{fig:research-progress}.
Stronger harnesses typically make sustained progress over the experiment
budget, whereas weaker methods plateau earlier or improve less consistently.
For example, TPE stops improving relatively early in the parameter search,
while GEAR and AutoScientists continue to find improvements near the end of
their recorded searches.
These trajectories provide complementary evidence that the benchmark scores
reflect systematic differences in research progress rather than isolated final
outcomes.

\noindent\textbf{Similar outcomes can hide different research efficiency.}
Endpoint performance does not fully characterize how a result is reached.
On the Opus~5 training-time objective, GEAR, Arbor, Beam Search, and Heuresis
all reach 270 seconds, but first attain this value after 12, 39, 50, and 99
charged experiments, respectively.
Thus, harnesses with identical final scores can differ substantially in how
efficiently they discover the same level of improvement.

\vspace{-2mm}
\subsection{A Broader Idea Frontier}
\label{sec:idea-frontier}
\vspace{-2mm}

\noindent\textbf{Multi-target research substantially broadens the idea frontier.}
Beyond routine hyperparameter tuning, Karpathy's reported loss-driven search
identifies two architectural changes, \emph{smear} and \emph{backout}
~\citep{karpathyautoresearch}.
Across Arena's efficiency objectives, our search catalogue contains 32 distinct
findings, including 20 that were accepted as improvements during search, which
we consolidate into nine idea families spanning model architecture, training,
and inference.
Different objectives expose qualitatively different research directions, from
architectural changes under model-efficiency targets to memory and systems
optimizations under training and inference targets.

\begin{wraptable}[25]{r}{0.42\textwidth}
\centering
\definecolor{FrontierBlue}{HTML}{195E87}
\definecolor{FrontierMuted}{HTML}{505963}
\definecolor{FrontierAfter}{HTML}{F0F5F8}
\definecolor{FrontierRule}{HTML}{C6CDD2}
\newcommand{\FrontierChange}[1]{{\color{FrontierBlue}#1}}
\newcommand{\FrontierLabel}[1]{%
  {\sffamily\fontsize{6.5}{7.5}\selectfont\bfseries
   \color{FrontierMuted}#1}}
\ifcsname hypersetup\endcsname\hypersetup{hidelinks}\fi
\captionsetup{font=footnotesize,labelfont=bf,skip=5pt,
  justification=raggedright,singlelinecheck=false}
\vspace{-0.18in}
\caption{A representative searched idea by improving parameter efficiency. Blue highlights searched changes; more discoveries are shown in Appendix~\ref{app:discovery-analysis}.}
\label{tab:loop-depth-aware}
\fontsize{8}{9.5}\selectfont
\setlength{\parindent}{0pt}
\setlength{\parskip}{0pt}
\setlength{\tabcolsep}{0pt}
\setlength{\jot}{0.5pt}
\renewcommand{\arraystretch}{0.65}
\arrayrulecolor{FrontierRule}
\begin{tabular}{@{}>{\raggedright\arraybackslash}p{\linewidth}@{}}
\hline
\addlinespace[4pt]
\textbf{Discovered idea:}
Loop Transformer with depth-specific residual gain\par
\vspace{3pt}
\textbf{Optimization target:} Parameter efficiency\\
\addlinespace[4pt]
\FrontierLabel{BEFORE: INDEPENDENT BLOCKS}\\
\hspace*{3pt}$\begin{aligned}[t]
&\mathbf{for}\ \ell=0,\ldots,7: \\
&\quad u \leftarrow h+\operatorname{Attn}_{\ell}(h)\\
&\quad h \leftarrow u+\operatorname{MLP}_{\ell}(u)
\end{aligned}$\\
\addlinespace[4pt]
\rowcolor{FrontierAfter}[0pt][0pt]
\FrontierLabel{AFTER: SHARED BLOCKS, DEPTH-AWARE GAINS}\\
\rowcolor{FrontierAfter}[0pt][0pt]
\hspace*{3pt}$\begin{aligned}[t]
&\mathbf{for}\ \ell=0,\ldots,7: \\
&\quad u \leftarrow h+
  \FrontierChange{\alpha_{\ell}\operatorname{Attn}_{\ell\bmod B}(h)}\\
&\quad h \leftarrow u+
  \FrontierChange{\beta_{\ell}\operatorname{MLP}_{\ell\bmod B}(u)}
\end{aligned}$\\
\addlinespace[5pt]
$B=1$: one stored block, eight applications.
The learned $\alpha_\ell,\beta_\ell$ control attention and MLP
residual updates separately at each depth.\\
\addlinespace[5pt]
\FrontierLabel{RELATION TO PRIOR WORK}\par\vspace{1pt}
Recent work scales residual updates with loop count
\citep{wang2026residual}; here the gains are learned separately
for each depth.\\
\addlinespace[4pt]
\bottomrule
\end{tabular}
\arrayrulecolor{black}
\par%
{\scriptsize\raggedright
\textbf{Notation.}\enspace
$h$: hidden state; $\ell$: depth; $B$: stored blocks.
Schematic: normalization and auxiliary paths omitted.\par}
\vspace{0.1in}
\end{wraptable}

\noindent\textbf{The discovered idea frontier mirrors active LLM research.}
The breadth of Arena's discoveries is not merely a larger collection of
benchmark-specific optimizations: many of them reach questions central to
current LLM research.
A representative example emerges from parameter efficiency
(Tab.~\ref{tab:loop-depth-aware}).
Agents first discover recurrent weight sharing, effectively turning the model
into a looped Transformer, and further identify the need to adapt residual
updates across repeated depth, learning separate attention and MLP residual
gains at each depth.
This closely mirrors recent work studying residual adaptation as a key
component of looped Transformers~\citep{wang2026residual}.
Similar correspondences appear across attention-state reduction through
cross-layer K/V sharing~\citep{brandon2024cla}, MQA/MLA~\citep{mqa,DeepSeek-V2},
conditional memory~\citep{cheng2026engram}, optimizer-state
reduction~\citep{zhang2025adammini}, activation-memory
efficiency~\citep{korthikanti2022recomputation}, and inference
optimization~\citep{aminabadi2022inference}.
Importantly, these discoveries are not simply reproductions of existing
methods: they often share the same underlying principles while arriving at
different technical mechanisms, offering alternative perspectives on active
research questions.
The resulting idea frontier therefore demonstrates a broader value of Arena:
its diverse objectives expand not only what research systems are evaluated on,
but also the space of meaningful research directions they explore.

\noindent\textbf{The broader idea frontier accompanies substantial measured progress.}
The searches producing these discoveries also achieve large admissible gains
across distinct resource dimensions.
In the Opus~5 group, the best programs reduce model size from 50.33M to 13.20M
parameters, training computation from 239.08M to 52.15M FLOPs per token, and
peak training memory from 43.96 to 0.373\,GiB, while satisfying BPB $<1.05$
and the corresponding resource constraints.
However, these endpoint programs combine multiple changes accumulated during
search, so their gains do not establish whether the discovered ideas themselves
provide reusable knowledge beyond the searches that found them.
We test this directly in Sec.~\ref{sec:discovery-validation}.

\begin{figure}[t]
  \centering
  \includegraphics[width=\linewidth]{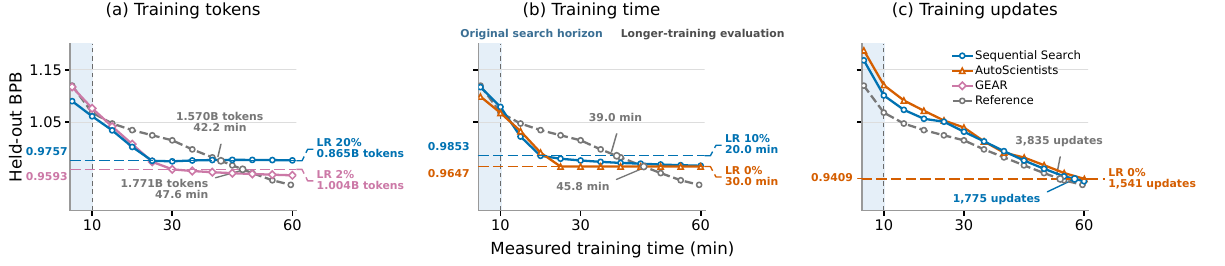}
  \vspace{-0.29in}
  \caption{\small{\textbf{Discovered programs can transfer beyond the search horizon.}
Programs discovered with 10-minute candidate training are retrained from random
initialization for 60 minutes alongside the reference.
Several retain their advantages for substantially longer than the original
search horizon, although these gains can diminish later as horizon-specific
training policies take effect.
Curves show held-out BPB against training time. Annotated comparisons use checkpoints after the learning-rate schedule reaches its floor, except in figure (c), where AutoScientists’ final BPB defines the shared target.}}
  \label{fig:scale-up}
  \vspace{-0.05in}
\end{figure}

\vspace{-2mm}
\subsection{Validating Discoveries Beyond the Search Setting}
\label{sec:discovery-validation}
\vspace{-2mm}

Arena discovers mechanisms under specific objectives and short experimental budgets. To assess their usefulness beyond these conditions, we first test whether a library of discovered ideas can guide a new search for better model quality. We then ask a stricter question: can winning programs from the original Arena searches retain their advantages when retrained for longer without further search?

\noindent\textbf{Combining discovered ideas improves model quality.}
We collect discoveries across Arena's objectives and harnesses into an idea library to guide a quality-driven search. The agent can select, modify, and combine mechanisms from different research trajectories rather than continue from a single winning program. We extend the training time budget to 1 hour in this search.

The resulting design improves model quality under both matched-budget evaluations (Fig.~\ref{fig:teaser}). At matched training data, it reduces training loss from 2.675 to 2.544 (4.9\%); at matched training FLOPs, it reduces training loss from 2.617 to 2.544 (2.8\%). Final held-out BPB also decreases from 0.9295 to 0.9112 at the same model size (50.33M parameters). The resulting design selectively adapts and simplifies ideas rather than merely combining their original implementations. These results demonstrate that efficiency-targeted discoveries can serve as reusable model-design knowledge for improving quality. Appendix~\ref{app:idea-reuse} details how these ideas are selected and transformed.

\noindent\textbf{Discovered programs can transfer far beyond the search horizon.}
We next ask a stricter question: can a program found under a short training
budget be scaled directly, without further search?
Arena normally evaluates candidate programs after 10 minutes of training.
We select two programs from each of the training-time, token, and update
objectives and retrain them from random initialization for 60 minutes alongside
the reference program.

Several discoveries retain their advantages far beyond the horizon at which
they were found.
A token-optimized GEAR program reaches 0.9593 BPB after 30 minutes using
1.004B tokens, while the reference requires approximately 47.6 minutes and
1.771B tokens to reach the same quality, corresponding to 43.3\% fewer tokens.
Its quality advantage persists until approximately 51 minutes, extending to
five times the original search horizon.
Similarly, a training-time program discovered by AutoScientists outperforms the
reference over the 15--46-minute interval and reaches 0.9647 BPB after
30 minutes, a quality level that the reference reaches only after approximately
45.8 minutes, corresponding to 34.5\% less training time
(Figure~\ref{fig:scale-up}; Table~\ref{tab:scale-up}).
Reference costs at matched quality are estimated by linear interpolation.

The transfer is substantial but not indefinite.
The GEAR program's learning rate reaches its 2\% floor around 27 minutes, while
the AutoScientists schedule reaches zero around 24 minutes.
Their early advantages consequently diminish as training continues; by
60 minutes, for example, the reference reaches 0.9295 BPB compared with
0.9478 for the GEAR program.
Thus, short-horizon search can discover programs whose advantages persist for
four to five times the original search horizon, while horizon-dependent
training policies can ultimately limit direct extrapolation to
longer training.

\subsection{The Science of Auto-Research}
\label{sec:science}

Arena also enables controlled study of what shapes the auto-research process. We examine quality constraints, budget allocation between evaluation depth and search breadth, and the role of research history in cross-target search.

\noindent\textbf{Constraints shape the research space.}
We vary the BPB threshold across $1.02$, $1.05$, and $2.0$ for parameter- and memory-efficiency searches with Sequential Search and Beam Search (Tab.~\ref{tab:quality-threshold-study}). Relaxing the threshold from $1.05$ to $2.0$ allows Sequential Search to reduce parameters from 18.94M to 0.69M, while its separate peak-memory search reduces memory from 0.83\,GiB to 0.035\,GiB. The additional quality slack permits substantially more aggressive efficiency improvements. Conversely, tighter thresholds restrict which structural changes remain admissible. Quality constraints thus shape not only the final results, but also which regions of the design space can be developed.

\noindent\textbf{Budget allocation creates a depth--breadth trade-off.}
For parameter minimization under BPB $<2.0$, we vary candidate training time from 1,200 to 6,000 seconds under a shared 20 GPU-hour training budget, trading fewer, longer evaluations for more, shorter ones (Tab.~\ref{tab:training-budget}). The shortest setting evaluates 60 candidates and finds the smallest model; longer settings yield better BPB but more parameters. These outcomes illustrate the trade-off between evaluation depth and search breadth, with different allocations favoring different quality--efficiency balances.

\begin{table}[t]
  \centering
  \footnotesize
  \captionsetup{font=footnotesize,skip=4pt,justification=raggedright,singlelinecheck=false}
  \begin{minipage}[t]{0.45\linewidth}
    \vspace{0pt}

\centering
\footnotesize
\setlength{\tabcolsep}{2.5pt}
\renewcommand{\arraystretch}{1}
\captionsetup{skip=3pt}

\caption{\textbf{Direct transfer beyond the search horizon.}
Programs discovered with 10-minute training are retrained for up to
60 minutes without further search.
Resource savings compare each program’s 30-minute checkpoint with the reference at the same held-out BPB.}
\label{tab:scale-up}

\resizebox{\linewidth}{!}{%
\begin{tabular}{@{}llccc@{}}
  \toprule
  \textbf{Objective} &
  \textbf{Harness} &
  \shortstack{\textbf{Resource}\\\textbf{saving}} &
  \shortstack{\textbf{Quality}\\\textbf{advantage to}} &
  \shortstack{\textbf{60-min}\\\textbf{BPB}} \\
  \midrule

  \multicolumn{2}{@{}l}{Reference}
    & -- & -- & 0.9295 \\
  \midrule

  \multirow{2}{*}{Training time}
    & Seq. Search & 28.5\% & $\sim$44 min & 0.9662 \\
    & AutoScientists & \textbf{34.5\%} & $\sim$46 min & 0.9647 \\
  \midrule

  \multirow{2}{*}{Training tokens}
    & GEAR & \textbf{43.3\%} & $\sim$51 min & 0.9478 \\
    & Seq. Search & 34.4\% & $\sim$42 min & 0.9765 \\
  \midrule

  \multirow{2}{*}{Training updates}
    & AutoScientists & 30.1\% & -- & 0.9409 \\
    & Seq. Search & 39.6\% & -- & 0.9365 \\

  \bottomrule
\end{tabular}%
}

  \end{minipage}\hfill
  \begin{minipage}[t]{0.50\linewidth}
    \vspace{0pt}
    
\centering
\scriptsize
\setlength{\tabcolsep}{0.5pt}
\renewcommand{\arraystretch}{1.28}
\caption{\small{\textbf{Looser quality constraints enable more aggressive efficiency gains.}
Results use 30 completed runs per search; bold marks each harness's best value.}}
\label{tab:quality-threshold-study}
\resizebox{\linewidth}{!}{%
\begin{tabular}{@{}llcccc@{}}
  \toprule
\multirow{2}{*}{\textbf{Harness}}
& \multirow{2}{*}{\shortstack{\textbf{BPB}\\\textbf{Limit}}}
& \multicolumn{2}{c}{\textbf{Parameters}}
& \multicolumn{2}{c}{\textbf{Peak GPU Memory}} \\
\cmidrule(lr){3-4}\cmidrule(lr){5-6}
& & Params. (M) & BPB & Memory (GiB) & BPB \\
\midrule
  \multirow{3}{*}{\shortstack[l]{Beam\\Search}}
    & $<1.02$ & 23.5934 & \textbf{1.0185} & 3.9753 & \textbf{1.0175} \\
    & $<1.05$ & 19.6610 & 1.0448 & 1.6707 & 1.0371 \\
    & $<2.0$ & \textbf{0.4649} & 1.5030 & \textbf{0.1625} & 1.2312 \\
  \midrule
  \multirow{3}{*}{\shortstack[l]{Sequential\\Search}}
    & $<1.02$ & 26.4767 & \textbf{1.0195} & 6.8889 & \textbf{1.0162} \\
    & $<1.05$ & 18.9399 & 1.0497 & 0.8341 & 1.0402 \\
    & $<2.0$ & \textbf{0.6924} & 1.9963 & \textbf{0.0348} & 1.7200 \\
  \bottomrule
\end{tabular}
\vspace{-0.1in}
}

  \end{minipage}
\end{table}

\newsavebox{\arenacrosstargettable}
\begin{table}[t]
  \centering
  \footnotesize
  \setlength{\tabcolsep}{15pt}
  \renewcommand{\arraystretch}{1.12}
  \caption{\small{
  \textbf{Research can build across objectives.}
  Each sequence continues from a program discovered under the first objective
  and optimizes a new objective.
  The second search can preserve the earlier gain or trade part of it for
  progress on the new objective.
  $\Delta_2$ reports the change from the first-search program; negative values
  indicate improvements.
  }}
  \vspace{-0.1in}
  \label{tab:cross-target-study}

  \begin{lrbox}{\arenacrosstargettable}
  \begin{tabular}{@{}lllcccc@{}}
    \toprule
    \textbf{Research sequence}
      & \textbf{Metric}
      & \textbf{Reference}
      & \shortstack{\textbf{After first}\\\textbf{search}}
      & \shortstack{\textbf{After second}\\\textbf{search}}
      & $\boldsymbol{\Delta_2}$ \\
    \midrule

    \multirow{2}{*}{Parameters $\rightarrow$ FLOPs}
      & Parameters (M)
      & 50.33 & 14.09 & 14.09 & $0.0\%$ \\
      & Training FLOPs (M/token)
      & 239.08 & 113.80 & 75.46 & $\mathbf{-33.7\%}$ \\

    \midrule

    \multirow{2}{*}{FLOPs $\rightarrow$ Decode latency}
      & Training FLOPs (M/token)
      & 239.08 & 66.55 & 66.55 & $0.0\%$ \\
      & Uncached decode latency (ms)
      & 552.09 & 520.36 & 65.17 & $\mathbf{-87.5\%}$ \\

    \midrule

    \multirow{2}{*}{Parameters $\rightarrow$ Training time}
      & Parameters (M)
      & 50.33 & 14.09 & 17.56 & $+24.6\%$ \\
      & Time to BPB $<1.05$ (s)
      & 600 & 570 & 450 & $\mathbf{-21.1\%}$ \\

    \bottomrule
  \end{tabular}
  \end{lrbox}

  \resizebox{
    \dimexpr\wd\arenacrosstargettable*
    \number\linewidth/\number\wd\arenamainresultstable\relax
  }{!}{\usebox{\arenacrosstargettable}}
  \vspace{-0.15in}
\end{table}

\noindent\textbf{Cross-target research builds on prior discoveries.}
We initialize searches under new objectives from programs discovered under earlier objectives, transferring code while reinitializing model weights (Tab.~\ref{tab:cross-target-study}). Subsequent searches can accumulate compatible gains: parameter savings are fully preserved while training FLOPs decrease by a further 33.7\%, and a FLOP-optimized program preserves its FLOP count while uncached decode latency decreases by 87.5\%. Cross-target continuation can also introduce trade-offs: optimizing training time from the parameter-optimized program reduces time to quality by 21.1\% but increases parameters by 24.6\%. Prior discoveries therefore provide foundations for both cumulative gains and new trade-offs across objectives.

Together, these studies show how constraints define admissible designs, budgets balance evaluation depth against search breadth, and research history supports cumulative gains or trade-offs across targets. Therefore, Arena enables systematic study of how autonomous research develops over time.

\vspace{-2mm}
\section{Conclusion}
\vspace{-2mm}

We introduced \methodname{}, a multi-target benchmark for autonomous research across nine quality-constrained language-model efficiency objectives. Arena reveals complementary harness strengths and a broad frontier of architectural and systems ideas. Discoveries can seed searches under new objectives, while a program combining selected ideas can improve model quality. Controlled studies further show how constraints, budgets, and research history shape progress. Together, these results support evaluating autonomous research across complementary objectives and studying how useful discoveries accumulate.

\noindent\textbf{Limitations.}
Our scope is explicitly limited to auto-research in language modeling, rather
than the full end-to-end research process or scientific novelty.
Idea-family groupings are interpretive, and noisy system measurements require
care when interpreting small differences.

\subsection*{AI use statement}
LLM-based research agents are the subjects of evaluation in this work; their roles in proposing ideas, implementing programs, and conducting experiments are described in the paper. 

Separately, we used AI tools to assist with literature search, compilation of experimental records, text polishing for readability and figure preparation. 

We have not used generative AI tools for helping develop theoretical models or conceptual frameworks, proposing or refining hypotheses, designing or providing feedback on research  methodology or experiments, implementing methods, assisting with translation, supporting qualitative and thematic data analysis, and interpreting results.

Generating synthetic data sets,  formulating mathematical claims, providing critical ingredients for proving mathematical claims, assisting in the writing of proofs, cleaning and reformating dataset, 
are not applicable to this work.

We have reviewed AI-assisted content against the relevant literature, source code, and experimental records. We take responsibility for the final content of this work, including text, claims or artifacts produced with the aid of generative AI.

\subsection*{Ethics statement}
This work studies autonomous research systems in a controlled language-modeling
environment. The experiments use existing language-model training data and do
not involve human subjects, personal data collection, or deployment to users.

The evaluated agents can autonomously propose code changes and execute
experiments, which may raise broader concerns about increasingly capable
autonomous research systems. In our setting, their actions are restricted to a
sandboxed experimental environment, a fixed program interface, predefined
compute budgets, and standardized evaluation. We do not evaluate or enable
autonomous interaction with external systems beyond the research environment.

We report both successful and unsuccessful research outcomes and distinguish
measured improvements from interpretations of scientific novelty. We hope that
the benchmark can support more systematic study of the capabilities and
limitations of autonomous research systems.

\bibliographystyle{iclr2027_conference}
\bibliography{iclr2027_conference}
\appendix

\clearpage
\hypersetup{hidelinks}
\section{Efficiency Target Definitions}\label{app:target-definitions}

The following specifications use the objective names in Table~\ref{tab:efficiency-targets}. Displayed resource limits are rounded; each data-pool constraint fixes the available pool rather than the number of tokens consumed.

\par\medskip\noindent\begin{minipage}{\linewidth}
\small\setlength{\parskip}{2pt}
\noindent\textbf{Total parameters}\par\nobreak

\textbf{Why it matters.} Fewer parameters shrink checkpoints and reduce model storage and transfer costs.

\textbf{How we measure it.} The model's total parameter count after the fixed 600-second training schedule.

        {\small\begin{tabularx}{\linewidth}{@{}p{.22\linewidth}X@{}}
\textbf{Optimize} & Minimize total parameters. \\
\textbf{Maintain quality} & Validation BPB \ensuremath{<} 1.05 (fixed quality threshold)  \\
\textbf{Stay within limits} & Training FLOPs \textbf{\ensuremath{\leq}239.1 M/token}\par Peak training memory \textbf{\ensuremath{\leq}44.0 GiB}\par Available training-data pool \textbf{Fixed 631.2 M-token pool} \\
\end{tabularx}}
\end{minipage}\par

\par\medskip\noindent\begin{minipage}{\linewidth}
\small\setlength{\parskip}{2pt}
\noindent\textbf{Training tokens to quality}\par\nobreak

\textbf{Why it matters.} Consuming fewer tokens reduces the data processing needed to train a model of the required quality.

\textbf{How we measure it.} Total training tokens consumed by the first scheduled validation check with BPB below 1.05, including warmup updates, gradient accumulation, and repeated use of data. This is token consumption, not the number of unique dataset positions visited.

        {\small\begin{tabularx}{\linewidth}{@{}p{.22\linewidth}X@{}}
\textbf{Optimize} & Minimize consumed training tokens. \\
\textbf{Maintain quality} & Validation BPB \ensuremath{<} 1.05 (fixed quality threshold)  \\
\textbf{Stay within limits} & Training FLOPs \textbf{\ensuremath{\leq}239.1 M/token}\par Total parameters \textbf{\ensuremath{\leq}50.3 M}\par Peak training memory \textbf{\ensuremath{\leq}44.0 GiB}\par Available training-data pool \textbf{Fixed 631.2 M-token pool} \\
\end{tabularx}}
\end{minipage}\par

\par\medskip\noindent\begin{minipage}{\linewidth}
\small\setlength{\parskip}{2pt}
\noindent\textbf{Training FLOPs}\par\nobreak

\textbf{Why it matters.} Fewer FLOPs reduce training computation independently of hardware and implementation speed.

\textbf{How we measure it.} Forward-and-backward FLOPs per token, measured on the uncompiled model using executed operations and registered counting formulas. The count includes matrix, convolution, and attention work, but excludes optimizer updates and unregistered operations such as elementwise operations and normalization.

        {\small\begin{tabularx}{\linewidth}{@{}p{.22\linewidth}X@{}}
\textbf{Optimize} & Minimize FLOPs per token. \\
\textbf{Maintain quality} & Validation BPB \ensuremath{<} 1.05 (fixed quality threshold)  \\
\textbf{Stay within limits} & Total parameters \textbf{\ensuremath{\leq}50.3 M}\par Peak training memory \textbf{\ensuremath{\leq}44.0 GiB}\par Available training-data pool \textbf{Fixed 631.2 M-token pool} \\
\end{tabularx}}
\end{minipage}\par

\par\medskip\noindent\begin{minipage}{\linewidth}
\small\setlength{\parskip}{2pt}
\noindent\textbf{Peak training memory}\par\nobreak

\textbf{Why it matters.} Lower memory lets training fit on devices with less capacity or use larger batches.

\textbf{How we measure it.} Peak PyTorch-allocated GPU memory from process setup through the reporting read, including FLOPs probing, training, and validation under the fixed 600-second schedule. Validation uses a fixed batch of 128 sequences of 2,048 tokens and can determine the peak. This is allocated memory, not allocator-reserved memory or total device usage.

        {\small\begin{tabularx}{\linewidth}{@{}p{.22\linewidth}X@{}}
\textbf{Optimize} & Minimize peak allocated memory. \\
\textbf{Maintain quality} & Validation BPB \ensuremath{<} 1.05 (fixed quality threshold)  \\
\textbf{Stay within limits} & Training FLOPs \textbf{\ensuremath{\leq}239.1 M/token}\par Total parameters \textbf{\ensuremath{\leq}50.3 M}\par Available training-data pool \textbf{Fixed 631.2 M-token pool} \\
\end{tabularx}}
\end{minipage}\par

\par\medskip\noindent\begin{minipage}{\linewidth}
\small\setlength{\parskip}{2pt}
\noindent\textbf{Training time}\par\nobreak

\textbf{Why it matters.} Shorter training time reaches the required quality sooner and captures hardware and implementation effects that FLOPs miss.

\textbf{How we measure it.} Measured training time at the first scheduled validation check with BPB below 1.05. Startup, compilation, evaluation, and the first ten warmup updates are excluded; each subsequent update includes all accumulation microbatches and the optimizer step, with GPU synchronization at its boundaries.

        {\small\begin{tabularx}{\linewidth}{@{}p{.22\linewidth}X@{}}
\textbf{Optimize} & Minimize measured training time. \\
\textbf{Maintain quality} & Validation BPB \ensuremath{<} 1.05 (fixed quality threshold)  \\
\textbf{Stay within limits} & Training FLOPs \textbf{\ensuremath{\leq}239.1 M/token}\par Total parameters \textbf{\ensuremath{\leq}50.3 M}\par Peak training memory \textbf{\ensuremath{\leq}44.0 GiB}\par Available training-data pool \textbf{Fixed 631.2 M-token pool} \\
\end{tabularx}}
\end{minipage}\par

\par\medskip\noindent\begin{minipage}{\linewidth}
\small\setlength{\parskip}{2pt}
\noindent\textbf{Training updates}\par\nobreak

\textbf{Why it matters.} Fewer updates indicate faster optimization progress, separate from the duration or arithmetic cost of each update.

\textbf{How we measure it.} Total logical optimizer updates completed by the first scheduled validation check with BPB below 1.05, including warmup updates. Accumulation microbatches within one update are not counted as separate updates.

        {\small\begin{tabularx}{\linewidth}{@{}p{.22\linewidth}X@{}}
\textbf{Optimize} & Minimize optimizer updates. \\
\textbf{Maintain quality} & Validation BPB \ensuremath{<} 1.05 (fixed quality threshold)  \\
\textbf{Stay within limits} & Training FLOPs \textbf{\ensuremath{\leq}239.1 M/token}\par Total parameters \textbf{\ensuremath{\leq}50.3 M}\par Peak training memory \textbf{\ensuremath{\leq}44.0 GiB}\par Available training-data pool \textbf{Fixed 631.2 M-token pool} \\
\end{tabularx}}
\end{minipage}\par

\par\medskip\noindent\begin{minipage}{\linewidth}
\small\setlength{\parskip}{2pt}
\noindent\textbf{Live-state memory (KV cache)}\par\nobreak

\textbf{Why it matters.} Retained state grows with the live context and limits either context length or concurrent requests during generation.

\textbf{How we measure it.} The increase in PyTorch-allocated GPU memory while retaining one request state after a 1,536-token prefill and 512 decoding steps, with capacity 2,048. Measurement follows a discarded warmup request and constructs the state with CUDA graph capture disabled. It measures retained allocation rather than transient peak memory or a logical tensor-size sum, and includes allocator rounding (approximately 1 MiB in the reference measurement).

        {\small\begin{tabularx}{\linewidth}{@{}p{.22\linewidth}X@{}}
\textbf{Optimize} & Minimize long-prompt retained GPU allocation. \\
\textbf{Maintain quality} & Validation BPB \ensuremath{<} 1.05 (fixed quality threshold)  \\
\textbf{Stay within limits} & Request latency \textbf{\ensuremath{\leq}750 ms}\par Training FLOPs \textbf{\ensuremath{\leq}239.1 M/token}\par Total parameters \textbf{\ensuremath{\leq}50.3 M}\par Peak training memory \textbf{\ensuremath{\leq}44.0 GiB}\par Available training-data pool \textbf{Fixed 631.2 M-token pool} \\
\end{tabularx}}
\end{minipage}\par

\par\medskip\noindent\begin{minipage}{\linewidth}
\small\setlength{\parskip}{2pt}
\noindent\textbf{Decode latency}\par\nobreak

\textbf{Why it matters.} Lower decode latency produces generated tokens faster after the prompt has been processed.

\textbf{How we measure it.} Median latency for a one-token prefill followed by 512 decoding steps, measured over 30 timed passes after three warmups. The timer includes prefill and synchronized GPU completion, but excludes state allocation and reset.

        {\small\begin{tabularx}{\linewidth}{@{}p{.22\linewidth}X@{}}
\textbf{Optimize} & Minimize latency. \\
\textbf{Maintain quality} & Validation BPB \ensuremath{<} 1.05 (fixed quality threshold)  \\
\textbf{Stay within limits} & Retained GPU state allocation at 513 tokens \textbf{\ensuremath{\leq}10 MiB}\par Training FLOPs \textbf{\ensuremath{\leq}239.1 M/token}\par Total parameters \textbf{\ensuremath{\leq}50.3 M}\par Peak training memory \textbf{\ensuremath{\leq}44.0 GiB}\par Available training-data pool \textbf{Fixed 631.2 M-token pool} \\
\end{tabularx}}
\end{minipage}\par

\par\medskip\noindent\begin{minipage}{\linewidth}
\small\setlength{\parskip}{2pt}
\noindent\textbf{Long-prompt latency (request latency)}\par\nobreak

\textbf{Why it matters.} Lower request latency reduces the full delay of processing a long context and generating a response.

\textbf{How we measure it.} Median latency for a 1,536-token prefill followed by 512 decoding steps, measured over 30 timed passes after three warmups. The timer includes prefill and synchronized GPU completion, but excludes state allocation and reset.

        {\small\begin{tabularx}{\linewidth}{@{}p{.22\linewidth}X@{}}
\textbf{Optimize} & Minimize latency. \\
\textbf{Maintain quality} & Validation BPB \ensuremath{<} 1.05 (fixed quality threshold)  \\
\textbf{Stay within limits} & Retained GPU state allocation at 2,048 tokens \textbf{\ensuremath{\leq}40 MiB}\par Training FLOPs \textbf{\ensuremath{\leq}239.1 M/token}\par Total parameters \textbf{\ensuremath{\leq}50.3 M}\par Peak training memory \textbf{\ensuremath{\leq}44.0 GiB}\par Available training-data pool \textbf{Fixed 631.2 M-token pool} \\
\end{tabularx}}
\end{minipage}\par
\par\medskip\noindent\textbf{Shared inference correctness requirements.}
The three inference tasks supply ground-truth tokens at each decoding step and compare the resulting next-token distributions with the same model's full forward pass. For both the one-token and 1,536-token prefill settings, the maximum total-variation distance across all 513 scored positions must not exceed 0.05. The returned request-state object must own all retained KV data, including compressed representations or replacement token history, and these data must remain on the GPU throughout the request. Memory constraints use the same retained-allocation procedure as the KV-cache objective, with capacities of 513 and 2,048 for the short- and long-prompt settings, respectively.

\section{Harness Integration and Reference Baselines}\label{app:reference-baselines}

This section specifies the integration interface and the two reference search policies used in Section~\ref{sec:experiments}. Arena provides execution, measurement, and scoring; the baselines determine proposals and retention. Their selection rules are harness choices, not requirements imposed on other integrations.

\subsection{Interface and Recorded Artifacts}
\label{app:reproducibility}

\noindent\textbf{Minimal interface.}
A harness must inspect the provided workspace and task definition, modify only the permitted program, submit experiments through Arena, read standardized results, and choose subsequent candidates. Arena defines the research problem and evaluation environment while leaving the research strategy open: an integration may retain an incumbent, beam, population, hypothesis tree, or another search state.

\noindent\textbf{Recorded artifacts.}
Each run records the reference program, proposed ideas, candidate implementations, execution outcomes, measured metrics, admissibility decisions, and search lineage. The terminal score is computed from measured admissible outcomes; the accompanying record supports audits and post-hoc trace analysis. Charged-experiment accounting follows Section~\ref{app:execution-budget}, independently of harness-side bookkeeping.

\subsection{Initialization}

Arena prepares the environment and evaluates the reference program. The harness receives its source and measurements, the task's objective, BPB threshold, resource constraints, measurement procedures, experiment budget, and a workspace for candidate programs and research records.

Sequential Search starts from one incumbent and proposes one candidate per round. Beam Search retains up to four programs and proposes two modifications per retained program in each generation. A full generation therefore contains eight candidate experiments; selection occurs after all outcomes are available.

A persistent controller coordinates an ideator, executor, and reviewer. The implementation calls the reviewer an \texttt{evaluator}, distinct from Arena's evaluator. Agent sessions are fresh for each round or generation. Instructions, source code, task rules, and saved research history are supplied to each session, so continuation does not rely on conversational memory.

\subsection{Proposal Preparation}

The ideator inspects the current code and prior experimental history, then records a proposal, rationale, timestamp, and position in that history. Edits may change architecture, attention, optimization, data processing, or training within the permitted boundary. A proposal can coordinate several edits implementing one mechanism; Sequential Search also requests follow-up experiments that could separate their contributions.

The executor implements each proposal in an isolated copy of its parent program and checks the edits before Arena's benchmark evaluation.

\subsection{Execution and Budget Accounting}
\label{app:execution-budget}

Arena checks the submitted source, protected files, available launch budget, and hardware, then runs training and task-specific evaluation. It records source, measurements, execution status, and logs. Successful execution alone does not establish admissibility.

Sequential Search uses one worker; Beam Search uses four. Parallel submission does not alter accounting: each candidate is one experiment. A full eight-candidate generation needs eight available slots and consumes eight launches if all reach chargeable GPU work, although at most four run simultaneously. Local CPU checks do not replace candidate training through Arena.

The budget is 100 charged launches, including the reference evaluation. A recorded GPU-work-start marker determines charging. Once GPU work begins, crashes, out-of-memory errors, and task failures still consume budget. A resolved launch without the marker is not charged; an unresolved launch reserves a slot. A normally charged reference therefore leaves at most 99 candidate experiments. The harness reads Arena's remaining budget rather than counting its own rounds or generations. Failed experiments remain recorded.

\subsection{Review and Selection}

The reviewer compares the proposal, parent and candidate code, measurements, and logs, then records feedback. Both baselines exclude failed or constraint-violating candidates from further development. Other harnesses may continue refining them, but only admissible results contribute to Arena's score.

\noindent\textbf{Unchanged-target attempts.}
An efficiency improvement may increase BPB while remaining below the quality threshold. Both baselines also permit limited attempts to improve quality without changing the target. Target equality means exact equality of recorded values, regardless of the proposal's description. Such attempts consume the allowance even when rejected.

\noindent\textbf{Sequential Search.}
An admissible candidate replaces the incumbent when it strictly improves the target. At exactly equal target value, it may replace the incumbent while the allowance remains if it improves either the task-defined tie-break metric or BPB. Otherwise the incumbent is retained. The allowance starts at four unchanged-target attempts and resets only after an accepted strict target improvement; accepting a quality improvement alone does not reset it.

\noindent\textbf{Beam Search.}
Parents remain fixed throughout a generation. After every outcome is resolved, the reviewer pools parents and new candidates, removes exact program duplicates, and excludes failures, constraint violations, and missing required ranking measurements. Parents need not be replaced when their modifications fail.

Up to two unchanged-target attempts are permitted per generation only if the best retained target strictly improved during the preceding two completed generations. This allowance is determined before proposal generation. When permitted, a parent and modification with equal target are reduced to the admissible member with lower BPB. The remaining programs are ranked by target, ordered task tie-break metrics, and finally BPB, retaining up to four. Remaining ties favor lower complexity, lower resource use, and then an older program where evidence is available. A read-only helper computes identities, admissibility checks, and ranking values; the reviewer applies the selection policy.

\subsection{Persistence, Interruption, and Stopping}

Saved state retains programs, measurements, decisions, feedback, and the remaining unchanged-target allowance, together with Beam generation membership and outcomes. Rejected and failed experiments remain in the history. Continuation reconciles outstanding requests with Arena and completes pending review before new proposals; recovery itself is not a new experiment.

Sequential Search normally stops when the budget is exhausted. Beam Search also stops when the remaining budget cannot cover two candidates per retained program; it does not shrink the generation or select from a partially evaluated one. Both record the stopping reason and unused budget.

Arena independently ranks all admissible recorded results and reports the best program, which need not be the final harness-selected endpoint. Computing this result requires no additional training.

\section{Discovery Evidence and Supplementary Analysis}
\label{app:discoveries}
\label{app:discovery-analysis}
\subsection{Representative Discovery Catalogue}
\label{sec:idea_frontier}
\label{app:discovery-catalogue}

This catalogue presents nine representative findings, detailing their mechanisms, objectives, and relationships to prior work.

Table~\ref{tab:searched-ideas} gives mechanisms, objectives, and comparisons with prior work. These examples use the BPB thresholds specified in Table 7. Correspondences identify related research questions and technical differences, not necessarily exact rediscoveries. Endpoints combine multiple changes and do not isolate individual mechanisms' contributions.

\begingroup
\definecolor{IdeaBlue}{HTML}{195E87}
\definecolor{IdeaMuted}{HTML}{505963}
\definecolor{IdeaAfter}{HTML}{F0F5F8}
\definecolor{IdeaBand}{HTML}{E9EEF1}
\definecolor{IdeaRule}{HTML}{C6CDD2}
\newcommand{\IdeaChange}[1]{{\color{IdeaBlue}#1}}
\newcommand{\IdeaTitle}[4]{%
  {\sffamily\fontsize{8.3}{9.5}\selectfont\bfseries
    \textcolor{IdeaBlue}{(#1)}\enspace #2}\par
  \vspace{1pt}{\fontsize{7.6}{8.6}\selectfont #3}\par
  \vspace{1.5pt}{\sffamily\scriptsize\color{IdeaMuted}#4}}
\newcommand{\IdeaLabel}[1]{%
  {\sffamily\fontsize{6.5}{7.5}\selectfont\bfseries\color{IdeaMuted}#1}}
\newcommand{\IdeaMath}[1]{%
  \hspace*{3pt}\(\begin{aligned}[t]#1\end{aligned}\)}
\newcommand{\IdeaRelation}[1]{%
  {\sffamily\fontsize{6.5}{7.5}\selectfont\bfseries\color{IdeaMuted}RELATION TO PRIOR WORK}\par
  \vspace{1pt}#1}
\newcommand{\IdeaSection}[2]{%
  \rowcolor{IdeaBand}[0pt][0pt]
  \multicolumn{3}{@{}l@{}}{%
    \hspace{4pt}\rule[-2pt]{0pt}{11pt}%
    {\sffamily\fontsize{7.3}{8.5}\selectfont\bfseries #1}%
    \quad{\sffamily\fontsize{7}{8.5}\selectfont\color{IdeaMuted}#2}}\\}
\ifcsname hypersetup\endcsname\hypersetup{hidelinks}\fi
\captionsetup{font=small,labelfont=bf,justification=raggedright,singlelinecheck=false,skip=3pt}
\begin{table}[p]
\centering
\fontsize{8}{9.5}\selectfont
\setlength{\jot}{0.5pt}
\setlength{\parindent}{0pt}\setlength{\parskip}{0pt}
\caption{Nine findings from Auto-Research Arena:
\textbf{six research findings} (a--f) and
\textbf{three infrastructure engineering findings} (g--i).
The latter address low-precision updates, activation memory,
and decode execution in large-model infrastructure.
Blue highlights searched changes.}
\label{tab:searched-ideas}
\setlength{\tabcolsep}{6pt}
\renewcommand{\arraystretch}{0.65}
\arrayrulecolor{IdeaRule}
\begin{tabularx}{\linewidth}{@{}>{\raggedright\arraybackslash}X
  >{\raggedright\arraybackslash}X >{\raggedright\arraybackslash}X@{}}
\IdeaSection{RESEARCH FINDINGS}{Model and optimizer design \enspace (a--f)}
\addlinespace[3pt]
\IdeaTitle{a}{Looped Transformers}{Depth-specific residual gains}{Parameters}
&
\IdeaTitle{b}{Lookup-augmented FFNs}{Token lookup before activation}{Training FLOPs \enspace $\operatorname{MLP}_\ell(h)$}
&
\IdeaTitle{c}{Cross-layer K/V sharing}{Layer-specific queries}{Live-state memory}
\\
\addlinespace[1pt]
\IdeaLabel{BEFORE} & \IdeaLabel{BEFORE} & \IdeaLabel{BEFORE} \\
\IdeaMath{&\textbf{for }\ell=0,\ldots,7:\\
    &\quad u \gets h+\operatorname{Attn}_\ell(h)\\
    &\quad h \gets u+\operatorname{MLP}_\ell(u)}
&
\IdeaMath{&z \gets W_\ell^{\mathrm{up}}h\\
    &\textbf{return }W_\ell^{\mathrm{down}}\phi(z)}
&
\IdeaMath{\mathbf K,\mathbf V &\gets \operatorname{KV}_\ell(h)\\
    y &\gets \operatorname{Attn}_\ell(h;\mathbf K,\mathbf V)}
\\
\addlinespace[1pt]
\rowcolor{IdeaAfter}[0pt][0pt]
\IdeaLabel{AFTER} & \IdeaLabel{AFTER} & \IdeaLabel{AFTER} \\
\rowcolor{IdeaAfter}[0pt][0pt]
\IdeaMath{&\textbf{for }\ell=0,\ldots,7:\\
    &\quad u \gets h+\IdeaChange{\alpha_\ell\operatorname{Attn}_{\ell\bmod B}(h)}\\
    &\quad h \gets u+\IdeaChange{\beta_\ell\operatorname{MLP}_{\ell\bmod B}(u)}}
&
\IdeaMath{&\IdeaChange{e \gets \operatorname{Gate}_\ell(h)\odot E_\ell[\tau]}\\
    &z \gets \IdeaChange{G_\ell^{\mathrm{up}}h+e}\\
    &\textbf{return }\IdeaChange{G_\ell^{\mathrm{down}}}\phi(z)}
&
\IdeaMath{&\IdeaChange{\textbf{if }\ell\in\mathcal P:}\\
    &\quad\mathbf K,\mathbf V\gets\operatorname{KV}_\ell(h)\\
    &y \gets \operatorname{Attn}_\ell(h;\IdeaChange{\mathbf K,\mathbf V})}
\\
\addlinespace[3pt]
\IdeaRelation{One block ($B=1$), eight uses; each depth learns residual gains, unlike loop-count scaling \citep{wang2026residual}.}
&
\IdeaRelation{Grouped FFNs receive context-gated token-ID features before activation; Engram \citep{cheng2026engram} instead retrieves hashed $N$-grams.}
&
\IdeaRelation{Unequal CLA-style groups \citep{brandon2024cla} share K/V with private queries; LISA \citep{mu2024lisa} instead shares and corrects attention weights.}
\\
\addlinespace[3.5pt]\midrule\addlinespace[3.5pt]
\IdeaTitle{d}{Low-memory Adam}{Token-row second moments}{Training memory}
&
\IdeaTitle{e}{Shared low-rank bases}{Shared reads; private projections}{Parameters}
&
\IdeaTitle{f}{Unified vocabulary}{Input / value / output}{Parameters}
\\
\addlinespace[1pt]
\IdeaLabel{BEFORE} & \IdeaLabel{BEFORE} & \IdeaLabel{BEFORE} \\
\IdeaMath{m_{ij} &\gets \operatorname{EMA}_1(g_{ij})\\
    v_{ij} &\gets \operatorname{EMA}_2(g_{ij}^{\,2})}
&
\IdeaMath{&\textbf{for }p\in\{q,k,v\}:\\
    &\quad y_p\gets L_p R_p h}
&
\IdeaMath{h_0 &\gets E^{\mathrm{in}}[\tau]\\
    v_\ell^{\mathrm{tok}} &\gets E_\ell^{\mathrm{val}}[\tau]\\
    o &\gets E^{\mathrm{out}}h}
\\
\addlinespace[1pt]
\rowcolor{IdeaAfter}[0pt][0pt]
\IdeaLabel{AFTER} & \IdeaLabel{AFTER} & \IdeaLabel{AFTER} \\
\rowcolor{IdeaAfter}[0pt][0pt]
\IdeaMath{m_{ij} &\gets \operatorname{EMA}_1(g_{ij})\\
    \IdeaChange{v_i} &\gets
      \IdeaChange{\operatorname{EMA}_2\bigl(d^{-1}\textstyle\sum_j g_{ij}^{\,2}\bigr)}}
&
\IdeaMath{&\textbf{for }p\in\{q,k,v\}:\\
    &\quad y_p\gets L_p\IdeaChange{R}h}
&
\IdeaMath{h_0 &\gets \IdeaChange{E}[\tau]\\
    v_\ell^{\mathrm{tok}} &\gets \IdeaChange{E}[\tau]\\
    o &\gets \IdeaChange{E}h}
\\
\addlinespace[3pt]
\IdeaRelation{Adam-mini-style \citep{zhang2025adammini} token-row second moments apply to input/value embeddings; first moments and output-head moments remain full.}
&
\IdeaRelation{Learned Q/K/V read basis, private output factors; ASVD \citep{yuan2023asvd} instead factorizes pretrained weights using activation statistics.}
&
\IdeaRelation{Input--output tying \citep{press2016tying} extends to token-indexed value residuals: one table for all three roles.}
\\
\addlinespace[4pt]
\IdeaSection{INFRASTRUCTURE ENGINEERING}{Training and inference execution \enspace (g--i)}
\addlinespace[3pt]
\IdeaTitle{g}{Stochastic rounding}{BF16 parameter write-back}{Update precision}
&
\IdeaTitle{h}{Activation reconstruction}{Backward from retained outputs}{Activation memory}
&
\IdeaTitle{i}{Decode compilation regions}{Gathers inside Q/K/V preparation}{Decode latency}
\\
\addlinespace[1pt]
\IdeaLabel{BEFORE} & \IdeaLabel{BEFORE} & \IdeaLabel{BEFORE} \\
\IdeaMath{w &\gets \theta+\Delta\theta\\
    \theta &\gets \operatorname{RN}_{\mathrm{BF16}}(w)}
&
\IdeaMath{&r \gets \operatorname{ReLU}(z)\\
    &s \gets r^2;\quad\operatorname{save}(r)\\
    &g_z \gets 2r\odot g_s}
&
\IdeaMath{e &\gets E[\tau]\\
    c,s &\gets T[t]\\
    q,k,v &\gets\operatorname{In}(h,e,c,s)}
\\
\addlinespace[1pt]
\rowcolor{IdeaAfter}[0pt][0pt]
\IdeaLabel{AFTER} & \IdeaLabel{AFTER} & \IdeaLabel{AFTER} \\
\rowcolor{IdeaAfter}[0pt][0pt]
\IdeaMath{w &\gets \theta+\Delta\theta\\
    \theta &\gets \IdeaChange{\operatorname{SR}_{\mathrm{BF16}}(w)}}
&
\IdeaMath{&s \gets \phi(z);\quad\IdeaChange{\operatorname{save}(s)}\\
    &g_z \gets \IdeaChange{2\sqrt{s}}\odot g_s}
&
\IdeaMath{F &\gets\IdeaChange{\operatorname{Compile}(\operatorname{Prep})}\\
    q,k,v &\gets F(h,\tau,t)}
\\
\addlinespace[3pt]
\IdeaRelation{BF16 storage is retained; stochastic write-back preserves small updates in expectation \citep{zamirai2020bfloat16}.}
&
\IdeaRelation{Reuse retained outputs for backward; selective recomputation \citep{korthikanti2022recomputation} instead reruns operations to save memory.}
&
\IdeaRelation{Execution-overhead optimization, as in DeepSpeed Inference \citep{aminabadi2022inference}; the baseline already has CUDA graphs.}
\\

\addlinespace[2pt]
\bottomrule
\end{tabularx}
\arrayrulecolor{black}
\vspace{3pt}
\begin{minipage}{\linewidth}
\scriptsize
\textbf{Scope.}
 (a,e) Beam Search and (f) Sequential Search, BPB $<2.0$; (b--d,g,h) Sequential Search and (i) Beam Search , BPB $<1.02$
\par\vspace{2pt}
\textbf{Notation.}\quad
$h$: hidden state; $\ell$: depth; $B$: stored blocks; $\tau$: token ID;
$\phi(z)=\operatorname{ReLU}(z)^2$; $\mathcal P$: K/V producer layers.
$\operatorname{Attn}_\ell$ and $\operatorname{MLP}_\ell$ are full modules;
supplied K/V retain layer-specific query/output projections.
$G$: grouped projections; $E_\ell$: table assigned to depth $\ell$;
$\operatorname{Gate}_\ell$: broadcast gate.
In (d), $g_{ij}$ is an embedding gradient with row width $d$;
$\operatorname{EMA}_{1,2}$ use Adam decay rates.
In (f), $E$ stores token vectors by row; $o$: logits.
In (g), $\Delta\theta$: computed update; RN/SR: nearest/stochastic rounding.
In (h), $g_x=\partial\mathcal L/\partial x$.
In (i), $T[t]$: rotary factors at position $t$;
$\operatorname{Prep}$ includes both gathers and Q/K/V preparation $\operatorname{In}$.
Normalization, other auxiliary paths, and decode-state reconstruction are omitted.
\end{minipage}
\end{table}
\endgroup

\clearpage
\subsection{Supplementary Protocols for Research-Process Studies}
\label{app:ablation-details}

\subsubsection{Budget Allocation}

Four parameter-minimization searches use BPB $<2.0$ and candidate training caps of 1,200, 2,400, 4,800, and 6,000 seconds. Obtaining 60, 30, 15, and 12 completed candidates, respectively, gives 20 nominal GPU-hours per setting (Table~\ref{tab:training-budget}). The searches share starting code, a 379,260,514-token data pool, seed 42, A100 80GB hardware, tokenizer, and held-out evaluator. Beam width four, two proposals per parent, two retuning slots, and four GPU workers are fixed.

\par\medskip\noindent
\begin{minipage}{\linewidth}
  \centering
  \small
  \setlength{\tabcolsep}{5pt}
  \renewcommand{\arraystretch}{1.1}
  \captionof{table}{\textbf{Searching results with varying training time.}
  We compare Beam Search for parameter minimization with different training times per candidate under a fixed 20 GPU-hour budget.
  Shorter evaluations allow more candidates to be explored.
  The shortest setting finds the smallest model satisfying BPB $<2.0$, while longer evaluations achieve better BPB with larger models.}
  \label{tab:training-budget}
  \begin{tabular}{rrrr}
    \toprule
    Training time/candidate (s) & Completed candidates & Parameters & BPB \\
    \midrule
    1,200 & 60 & 123,272 & 1.986241 \\
    2,400 & 30 & 3,194,960 & 1.201839 \\
    4,800 & 15 & 11,542,672 & 1.060628 \\
    6,000 & 12 & 7,864,464 & 1.151816 \\
    \bottomrule
  \end{tabular}
\end{minipage}\par\medskip

These comparisons vary duration, candidate count, explored programs, and achieved quality together; they establish neither an optimal allocation nor a causal or matched-quality effect.

\subsubsection{Cross-Target Transfer Protocol}

Table~\ref{tab:cross-target-study} reports parameter-to-FLOP, FLOP-to-uncached-decode-latency, and parameter-to-training-time continuations. Program code is transferred; weights are freshly initialized. Each receiving evaluator remeasures the transferred program under its own objective and constraints. Earlier gains are not imposed as hard constraints.

There is one search per receiving objective. These selected examples demonstrate compatible improvements and permitted trade-offs; without matched restarts from the reference program, they establish neither a causal benefit of transfer nor a transfer-success rate.

\clearpage

\begin{figure}[t]
    \centering
    \includegraphics[width=\linewidth]
    {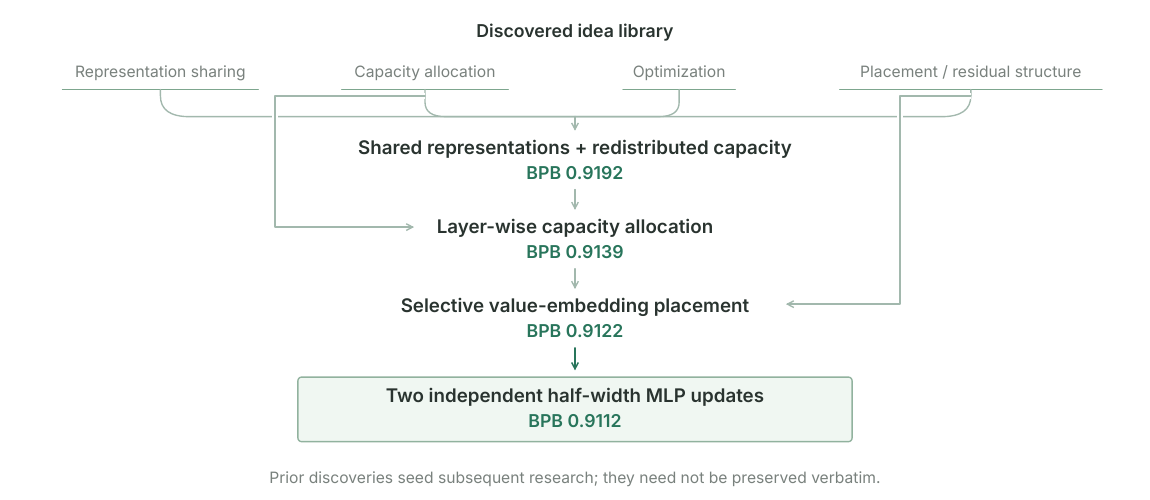}
    \vspace{-0.08in}
    \caption{\small{
    \textbf{The selected design emerges through reuse and reinterpretation of prior discoveries.}
    The quality-driven search builds on mechanisms for representation sharing
    and capacity allocation, subsequently changing where these mechanisms act
    and ultimately reinterpreting the MLP update as two independent half-width
    sublayers.
    Rather than simply accumulating prior discoveries, the search selects,
    recombines, and modifies them into new designs.
    }}
    \label{fig:idea-reuse-lineage}
    \vspace{-0.08in}
\end{figure}

\FloatBarrier
\subsection{How Discovered Ideas Are Reused}
\label{app:idea-reuse}

Section~\ref{sec:discovery-validation} shows that ideas discovered in Arena's
short, target-specific searches can be reused to improve model quality at a
larger training scale. Each candidate trains on a shared 10B-token pool for at most one hour or 2.217B tokens, whichever comes first.
Here we examine how this reuse occurs.
Rather than transferring a single winning program or mechanically combining all
previous discoveries, the quality-driven search receives an idea library and
can select, combine, modify, or discard its contents.
We analyze the provenance of the selected design, representative interactions
encountered during search, and how the candidate distribution evolves across
search rounds.

\noindent\textbf{The search draws on discoveries across objectives and harnesses.}
The recorded sources represented in the selected design span four research
harnesses and four classes of Arena objectives: parameter efficiency, training
tokens, training time, and training updates.
They include mechanisms for sharing representations, allocating MLP capacity,
changing optimization schedules, gating attention and value residuals,
restricting context, and changing where computations are applied across depth.
The resulting search is therefore not a continuation of one original objective
or one harness trajectory.
Instead, discoveries from heterogeneous searches form a common library of
candidate research ingredients.

\noindent\textbf{Reuse progressively transforms prior mechanisms.}
Figure~\ref{fig:idea-reuse-lineage} summarizes the main architectural
progression leading to the selected design.
An early improvement shares the value-embedding table and reallocates MLP
capacity, reducing BPB from the reference value of 0.9295 to 0.9192.
Subsequent search changes how capacity is distributed across depth and where
value embeddings are injected, eventually reaching 0.9122.
The final step changes a different aspect of the architecture: instead of one
MLP update per block, the model applies two independent half-width MLP
sublayers, reaching 0.9112 BPB.
Thus, the selected design is not a direct union of library entries.
The search progressively reinterprets how previously discovered mechanisms are
implemented and combined.

\begin{table*}[t]
\centering
\footnotesize
\setlength{\tabcolsep}{4.5pt}
\renewcommand{\arraystretch}{1.10}

\caption{\small{
\textbf{Recorded idea sources represented in the selected search lineage.}
The quality-driven search draws on mechanisms originally discovered by four
research harnesses under four classes of Arena objectives.
The provenance records indicate where an idea entered the library, not an
independent causal contribution to the final model.
}}
\label{tab:idea-reuse-sources}

\begin{tabular}{
    @{}
    p{0.22\textwidth}
    p{0.17\textwidth}
    p{0.16\textwidth}
    p{0.39\textwidth}
    @{}
}
\toprule
\textbf{Recorded idea}
&
\textbf{Source harness}
&
\textbf{Objective}
&
\textbf{Use in the quality-driven search}
\\
\midrule

Shared value-embedding table
&
Sequential Search
&
Parameters
&
Shares token-dependent value representations rather than maintaining separate
lookup tables.
\\

MLP width as a search variable
&
GEAR
&
Parameters
&
Reallocates model capacity through MLP width rather than uniformly shrinking
the architecture.
\\

Reduced weight decay
&
GEAR
&
Training tokens
&
Changes optimization while retaining the architectural backbone.
\\

Square-root cooldown
&
Heuresis
&
Training time
&
Changes the learning-rate schedule during later training.
\\

Per-head attention output gate
&
Sequential Search
&
Training time
&
Introduces learned control over individual attention-head outputs.
\\

Sparse value-residual gate channels
&
AutoScientists
&
Parameters
&
Restricts value-residual gating to a subset of channels.
\\

Shorter attention window
&
GEAR
&
Training updates
&
Changes the amount of context processed by attention.
\\

Layer-specific MLP widths
&
AutoScientists
&
Parameters
&
Redistributes MLP capacity across depth while approximately preserving total
width.
\\

Value embeddings at selected stack depths
&
Sequential Search
&
Training time
&
Moves value-residual injection layers at depth \{1,3,5,7\} to \{0,1,6,7\}.
\\

Two independent half-width MLP sublayers
&
Heuresis
&
Training tokens
&
Motivates a final design with two independent MLP updates within each block.
\\

\bottomrule
\end{tabular}
\end{table*}

\noindent\textbf{Idea provenance spans multiple research histories.}
Table~\ref{tab:idea-reuse-sources} lists the ten recorded idea sources
represented in the selected search lineage.
They originate from Sequential Search, GEAR, AutoScientists, and Heuresis, and
from searches targeting parameters, training tokens, training time, and
training updates.
This diversity supports the intended role of the idea library: mechanisms
discovered under one research objective can become hypotheses for subsequent
research under another.
At the same time, provenance only records where an idea entered the search;
it does not establish that every listed mechanism independently contributes to
the final improvement.

\begin{table}[t]
\centering
\footnotesize
\setlength{\tabcolsep}{4pt}
\renewcommand{\arraystretch}{1.10}

\caption{\small{
\textbf{Idea reuse is selective rather than additive.}
Representative outcomes from the quality-driven search show that plausible
combinations can help or hurt.
These comparisons arise from the search process and are not controlled
ablations.
}}
\label{tab:idea-reuse-cases}

\begin{tabular}{
    @{}
    p{0.73\linewidth}
    >{\raggedleft\arraybackslash}p{0.17\linewidth}
    @{}
}
\toprule
\textbf{Representative design}
&
\textbf{BPB $\downarrow$}
\\
\midrule

Fixed reference
&
0.929497
\\

Shared value embedding + redistributed MLP capacity
&
0.919165
\\

Reapply the same MLP without introducing independent capacity
&
0.934888
\\

Redistribute approximately the same total MLP width across depth
&
0.913908
\\

Inject value embeddings only at selected stack depths
&
0.912195
\\

Further tune the selective-injection design
&
0.911221
\\

Two independent half-width MLP sublayers
&
\textbf{0.911213}
\\

\bottomrule
\end{tabular}
\end{table}

\noindent\textbf{Useful ideas do not combine monotonically.}
Representative outcomes in Table~\ref{tab:idea-reuse-cases} reveal both
positive and negative interactions.
For example, simply reapplying an existing MLP adds another transformation
without introducing independent parameters and performs worse than the
reference, whereas introducing a genuinely independent second MLP appears in
the strongest final design.
Other search outcomes show that individually strong modifications can lose
their advantage when combined and that different allocations of approximately
the same total MLP width can yield different quality.
These observations are not controlled ablations, but they show why the idea
library should not be viewed as a collection of independently additive
improvements.
Its value instead lies in providing structured hypotheses that subsequent
research can test, reject, and reinterpret.

\begin{figure}[t]
    \centering
    \includegraphics[width=\linewidth]
    {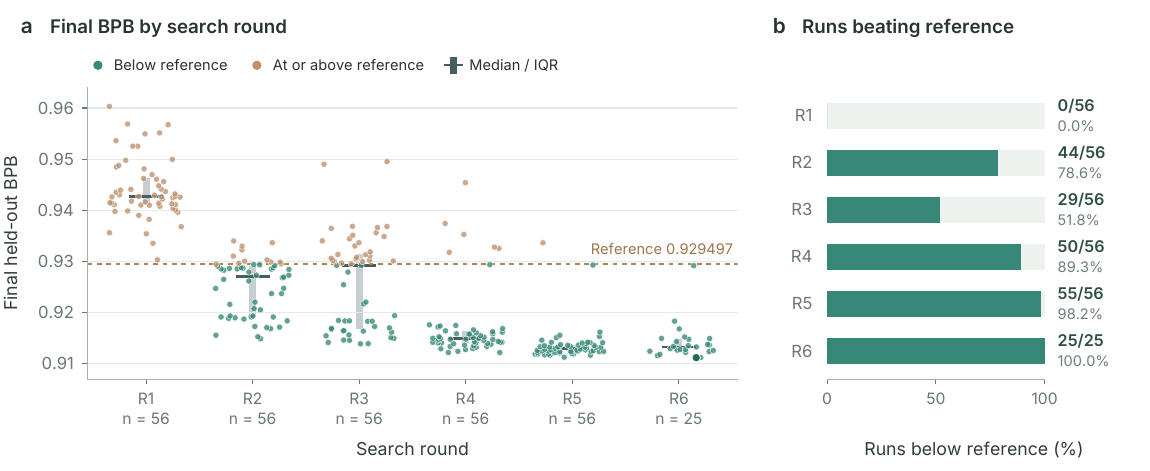}
    \vspace{-0.3in}
    \caption{\small{
    \textbf{The quality-driven search progressively moves into a stronger region.}
    Each point shows the final held-out BPB of one valid candidate, grouped by
    search round; horizontal jitter is used only to avoid overlap.
    Green points outperform the fixed reference (dashed line), while gray bars
    show the median and interquartile range within each round.
    The distribution shifts substantially below the reference after the first
    round and becomes increasingly concentrated at lower BPB in later rounds.
    Overall, 203 of 305 valid candidates outperform the reference.
    }}
    \label{fig:idea-reuse-search}
    \vspace{-0.08in}
\end{figure}

\noindent\textbf{Search progressively moves into a stronger design region.}
Figure~\ref{fig:idea-reuse-search} shows the distribution of candidate quality
across the six search rounds.
None of the 56 valid candidates in the first round outperforms the fixed
reference, compared with 44/56, 29/56, 50/56, 55/56, and 25/25 in subsequent
rounds.
The temporary broadening in the third round shows that the search does not
simply accumulate accepted improvements, while later rounds increasingly
concentrate around substantially lower BPB.
Across all rounds, 203 of 305 valid candidates outperform the reference,
indicating that the improvement reflects a broad region of stronger designs
rather than a single isolated candidate.

\noindent\textbf{The exact top-ranked combination should not be overinterpreted.}
Although the search moves toward stronger designs overall, its best candidates
are extremely close.
The two best recorded candidates achieve BPB values of 0.911213 and 0.911221,
differing by only $8\times10^{-6}$ in their one-hour evaluations.
With one measurement per candidate, differences at this scale do not establish
that either design is uniquely superior to nearby alternatives.
We therefore interpret the experiment as evidence that the discovered idea
library guides subsequent research toward a productive region of model designs,
rather than that one particular combination is uniquely optimal.

\section{Elo Scores}\label{app:elo}

Table~\ref{tab:main-results} reports the best recorded admissible target values across nine objectives for Opus~5 and GPT-6. We aggregate these values using the same margin-aware Elo procedure, with equal weight assigned to each objective.

Training times are compared after rounding to the nearest 30-second checkpoint; matching checkpoints are ties. Ratings are calculated from these stored target values before display rounding or unit conversion. The two agent-model groups use separate method pools, each with a mean rating of 1500, so Elo comparisons are made within each group.

All harnesses begin each calculation with the initial rating $R=1500$. Lower values are better for every included objective.

\par\medskip\noindent\textbf{Turn best admissible results into pairwise outcomes}\par\nobreak

      $$
      y_{a,b,j}=
      \begin{cases}
      1, & x_{a,j} < x_{b,j} \\
      \tfrac{1}{2}, & x_{a,j}=x_{b,j} \\
      0, & x_{a,j} > x_{b,j}
      \end{cases}
      $$

\emph{x}\textsubscript{\emph{a},\emph{j}} is harness \emph{a}'s best recorded admissible target value on objective \emph{j}. \emph{y}\textsubscript{\emph{a},\emph{b},\emph{j}} is the outcome when harness \emph{a} is compared with harness \emph{b}: 1 for a win, \ensuremath{\tfrac12} for a tie, and 0 for a loss.

\par\medskip\noindent\textbf{Estimate the expected outcome from the current ratings}\par\nobreak

      $$
      E_{a,b}=\frac{1}{1+10^{(R_b-R_a)/400}}
      $$

\emph{R}\textsubscript{\emph{a}} and \emph{R}\textsubscript{\emph{b}} are the harnesses' current ratings. \emph{E}\textsubscript{\emph{a},\emph{b}} is the outcome expected for harness \emph{a}, between 0 and 1. A higher-rated harness is expected to win more often.

\par\medskip\noindent\textbf{Measure how large the win is}\par\nobreak

      $$
      d_{a,b,j}=\log\!\left(\frac{x_{b,j}}{x_{a,j}}\right),
      \qquad
      z_{a,b,j}=\frac{|d_{a,b,j}|}{s_j},
      \qquad
      m_{a,b,j}=\min\!\left(3,\frac{\log(1+z_{a,b,j})}{\log 2}\right)
      $$

\emph{d}\textsubscript{\emph{a},\emph{b},\emph{j}} is the logarithm of the ratio between the two target values. Using a ratio makes proportional improvements comparable within an objective. \emph{s}\textsubscript{\emph{j}} is the median nonzero absolute log margin between all harness pairs on objective \emph{j}, bounded below by $10^{-9}$; it defaults to 1 when every pair is tied. It defines a typical margin for that objective. \emph{z}\textsubscript{\emph{a},\emph{b},\emph{j}} expresses the observed margin in units of that typical margin. The margin multiplier \emph{m}\textsubscript{\emph{a},\emph{b},\emph{j}} grows gradually with the margin and is capped at 3 so that one unusually large win cannot dominate the ranking. Exact ties have a zero multiplier and cause no rating update.

\par\medskip\noindent\textbf{Weight objectives and update both ratings}\par\nobreak

      $$
      \Delta R_a
      =K\,\frac{w_j}{\bar{w}}\,m_{a,b,j}\bigl(y_{a,b,j}-E_{a,b}\bigr),
      \qquad
      R_a\leftarrow R_a+\Delta R_a,
      \qquad
      R_b\leftarrow R_b-\Delta R_a
      $$

\emph{K} is 32, which controls the size of each update. \emph{w}\textsubscript{\emph{j}} is objective \emph{j}'s weight, and \emph{w\ensuremath{\bar{\phantom{w}}}} is the average objective weight. Dividing by the average keeps an ordinary comparison near the standard Elo update size. The term \emph{y} \ensuremath{-} \emph{E} rewards results that exceed the current expectation and penalizes results that fall short.

\par\medskip\noindent\textbf{Reduce sensitivity to comparison order}\par\nobreak

Because Elo updates happen one comparison at a time, their order can slightly affect the result. The calculator therefore generates 4,000 permutations using NumPy's default random generator with seed 20260902. Each pass starts all harnesses at 1500 and processes every pair once per objective; the reported score is the mean final rating across passes. The zero-sum updates preserve a cohort mean of 1500 for both agent-model groups, without a final shift or fixed maximum. The published score is a relative rating of the best recorded admissible results, not a percentage or a measurement in any resource unit.

\section{Extended Related Work}
\label{app:related-work}

\subsection{Autonomous-Research Benchmarks}

\noindent\textbf{Implementation, reproduction, and research tasks.}
Benchmarks cover repository execution, method implementation, reproduction, and expert-specified extensions~\citep{SUPER,CORE-Bench,ResearchCodeBench,LMR-Bench,PaperBench,EXP-Bench,RExBench}. Related capability suites assess software and command-line tasks~\citep{SWE-bench,Terminal-Bench}, retrieval~\citep{BrowseComp,AutoResearchBench}, scientific discovery and programming~\citep{DiscoveryBench,ScienceAgentBench,SciCode}, and theorem proving~\citep{PutnamBench}. Arena instead leaves the improvement mechanism open and evaluates a measured optimization target under quality and resource constraints.

\noindent\textbf{Open-ended research and research artifacts.}
Machine-learning engineering, research competitions, and paper-derived research environments support open-ended experimentation under task-specific evaluation protocols~\citep{MLE-Bench,BioML-Bench,RE-Bench,MLGym,ResearchGym,AIRS-Bench,InnovatorBench,MLRC-Bench}. Other evaluations examine proposals, experiments, reports, alignment, and artifact support across research stages~\citep{AI-Researcher,AutoResearchClaw,MLR-Bench,ResearchClawBench,ARAC-Bench,ResearchArena-AutoResearch,AstaBench}. Arena focuses on how harness performance changes across objectives in one language-modeling environment, ranking measured endpoints and using traces as explanatory evidence.

\noindent\textbf{Language-model and computational optimization.}
Language-model benchmarks evaluate validation loss, training speed, reproduction of historical improvements, or post-training under compute constraints~\citep{karpathyautoresearch,NanoGPT-Bench,Automated-LLM-Speedrunning,PostTrainBench}; systems benchmarks target kernels, numerical routines, and software performance under correctness requirements~\citep{KernelBench,AlgoTune,GSO}. Arena spans nine model, training, and inference objectives with a common quality requirement and objective-specific resource limits.

\noindent\textbf{Evaluation validity and reliability.}
Studies examine efficiency and scaffold contributions after benchmark saturation~\citep{CORE-Bench-Saturation}, specification and evaluator inconsistencies~\citep{BenchGuard}, and hardware sensitivity and score aggregation~\citep{Performance-Optimization-Benchmark-Audit}. These concerns motivate distinguishing valid resource improvements from apparent gains caused by evaluation artifacts.

\subsection{Autonomous-Research Harnesses}

\noindent\textbf{End-to-end workflows.}
Research systems connect literature review, idea generation, implementation, experiments, analysis, and reporting~\citep{lu2024aiscientist,AI-Researcher,AutoML-Agent,Agent-Laboratory}. The research harness orchestrates these activities around a foundation model.

\noindent\textbf{Tree, archive, and evolutionary search.}
Harnesses allocate experiments through code trees, hypothesis trees, archives, and alternative selection policies~\citep{AIDE,AIRA,Arbor,Heuresis}. Population-based methods retain competing programs or heuristics and generate candidates using executable feedback, mutation, or crossover~\citep{FunSearch,EoH,AlphaEvolve,GEAR}. These search procedures organize exploration; performance gains alone do not establish scientific novelty.

\noindent\textbf{Multi-agent coordination and persistent memory.}
Teams, proposal critique, execution recovery, and shared successful or failed experiments support sustained research~\citep{AutoScientists,AutoResearchClaw}. Semantic, episodic, and structured scientific memory provide complementary mechanisms for retaining knowledge across sessions~\citep{MLZero,Co-Scientist,Kosmos}.

\noindent\textbf{Harness design and self-improvement.}
Other methods optimize agent architectures, workflows, or implementations through meta-search and recursive improvement~\citep{ADAS,AFlow,Darwin-Godel-Machine,AIDE2}. Their search object is the research procedure itself, rather than only the candidate experimental program.

Arena evaluates harness--model combinations across multiple optimization objectives in a shared language-modeling environment, leaving the search strategy open and retaining standardized experimental traces.

\end{document}